\pdfoutput=1

\documentclass[11pt]{article}

\usepackage[final]{acl}

\usepackage{times}
\usepackage{latexsym}

\usepackage[T1]{fontenc}

\usepackage[utf8]{inputenc}

\usepackage{microtype}

\usepackage{inconsolata}

\usepackage{graphicx}

\usepackage{amsmath}

\DeclareMathOperator*{\argmin}{arg\,min}
\usepackage{multirow}
\usepackage{multicol}
\usepackage{makecell}
\usepackage{pifont}
\usepackage{amsfonts}
\usepackage{graphicx}
\usepackage{booktabs}
\usepackage{hyperref}

\title{PRIM: Towards Practical In-Image Multilingual Machine Translation}

\author{
    Yanzhi Tian\textsuperscript{1} ~~~ Zeming Liu\textsuperscript{2} ~~~ Zhengyang Liu\textsuperscript{1} ~~~ Chong Feng\textsuperscript{1} \\
    {\bf Xin Li}\textsuperscript{1} ~~~ {\bf Heyan Huang}\textsuperscript{1} ~~~ {\bf Yuhang Guo}\textsuperscript{1}\thanks{Corresponding author}
    \\
    \textsuperscript{1}School of Computer Science and Technology, Beijing Institute of Technology
    \\ 
    \textsuperscript{2}School of Computer Science and Engineering, Beihang University
    \\
    \texttt{\normalsize \{tianyanzhi,zhengyang,fengchong,xinli,hhy63,guoyuhang\}@bit.edu.cn;} \texttt{\normalsize zmliu@buaa.edu.cn}
    }

\begin{document}
\maketitle
\begin{abstract}
In-Image Machine Translation (IIMT) aims to translate images containing texts from one language to another. Current research of end-to-end IIMT mainly conducts on synthetic data, with simple background, single font, fixed text position, and bilingual translation, which can not fully reflect real world, causing a significant gap between the research and practical conditions. To facilitate research of IIMT in real-world scenarios, we explore Practical In-Image Multilingual Machine Translation (IIMMT). In order to convince the lack of publicly available data, we annotate the PRIM dataset, which contains real-world captured one-line text images with complex background, various fonts, diverse text positions, and supports multilingual translation directions. We propose an end-to-end model VisTrans to handle the challenge of practical conditions in PRIM, which processes visual text and background information in the image separately, ensuring the capability of multilingual translation while improving the visual quality. Experimental results indicate the VisTrans achieves a better translation quality and visual effect compared to other models. The code and dataset are available at: \url{https://github.com/BITHLP/PRIM}.
\end{abstract}

\section{Introduction}

In-Image Machine Translation (IIMT) aims to transform images containing texts from one language to another \cite{mansimov-etal-2020-e2eiimt, tian-etal-2023-pixelseq2seq, tian-etal-2025-debackx,lan-etal-2024-translatotronv, qian-etal-2024-anytrans}.
The challenge of IIMT lies in that both the input and output are images, detaching from the text modality, which is a significant distinction from other Neural Machine Translation (NMT) tasks incorporating image modality \cite{zhu-etal-2023-peit, liang-etal-2024-documentit, ma-etal-2024-e2etit, li-etal-2024-mit10m, zhang-etal-2025-layoutdit, fang-etal-2022-phraseMMT, chen-etal-2025-clearerMMT}, as they still center around text, with the input or output remaining text-based.

\begin{figure}[t]
    \centering
    \setlength{\tabcolsep}{3pt}
    \renewcommand{\arraystretch}{0.8}
    \begin{tabular}{cccccc}
    \Xhline{1.5pt}
    \multirow{2}{*}{\textbf{Name}} & \multirow{2}{*}{\textbf{Source}} & \multicolumn{4}{c}{\small \textbf{Features}} \\
    \cline{3-6}
    & & {\small \textbf{ML.}}& {\small \textbf{RB.}} & {\small \textbf{Fonts.}} & {\small \textbf{Pos.}}\\
    \hline
    {\small E2E-IIMT \shortcite{mansimov-etal-2020-e2eiimt}} & {\small Synth.} & \textcolor{red}{\ding{55}} & \textcolor{red}{\ding{55}} & \textcolor{red}{\ding{55}} & \textcolor{red}{\ding{55}} \\
    {\small SegPixel \shortcite{tian-etal-2023-pixelseq2seq}} & {\small Synth.} & \textcolor{red}{\ding{55}} & \textcolor{red}{\ding{55}} & \textcolor{green!70!black}{\ding{51}} & \textcolor{red}{\ding{55}} \\
    {\small TranslatotronV \shortcite{lan-etal-2024-translatotronv}} & {\small Synth.} & \textcolor{red}{\ding{55}} & \textcolor{red}{\ding{55}} & \textcolor{red}{\ding{55}} & \textcolor{green!70!black}{\ding{51}} \\
    {\small UMTIT \shortcite{niu-etal-2024-umtit}} & {\small Synth.} & \textcolor{red}{\ding{55}} & \textcolor{red}{\ding{55}} & \textcolor{red}{\ding{55}} & \textcolor{green!70!black}{\ding{51}} \\
    {\small DebackX \shortcite{tian-etal-2025-debackx}} & {\small Synth.} & \textcolor{red}{\ding{55}} & \textcolor{green!70!black}{\ding{51}} & \textcolor{green!70!black}{\ding{51}} & \textcolor{green!70!black}{\ding{51}} \\
    \hline
    {\small PRIM (Ours)} & {\small Real.} & \textcolor{green!70!black}{\ding{51}} & \textcolor{green!70!black}{\ding{51}} & \textcolor{green!70!black}{\ding{51}} & \textcolor{green!70!black}{\ding{51}} \\
    \Xhline{1.5pt}
    \end{tabular}
    
    \vspace{0.2em}
    
    \includegraphics[width=\linewidth]{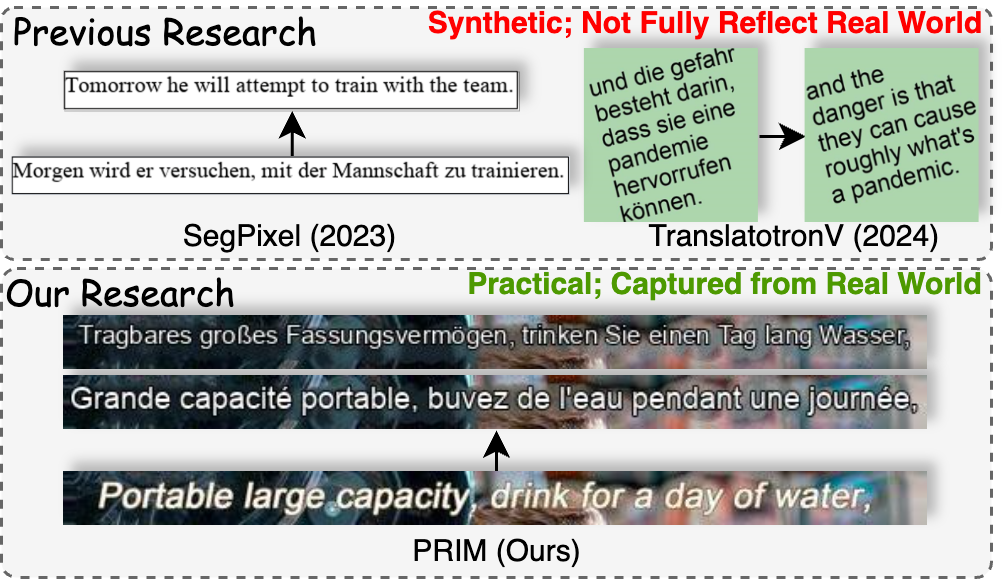}
    
    \caption{Comparison between PRIM to publicly available IIMT datasets. The sources of datasets are divided into synthetic (Synth.) and real world (Real.), and ``ML.'', ``RB.'', ``Fonts'' and ``Pos.'' specifies whether the dataset including multilingual translation, real-world backgrounds, various fonts and different text positions. 
    Previous research primarily conducts on synthetic data with simple background, single font, fixed text position, and bilingual translation. 
    Our research utilizes real-world captured images with complex background, various fonts, diverse text positions, and multilingual translation, which is more aligned to real conditions.}
    \label{fig:example}
\end{figure}

The translated target images of IIMT help people understand texts in visual modality directly, holding significant application value in translation software. 
A commonly used approach for IIMT is the cascade model: It begins with using an Optical Character Recognition (OCR) model to recognize text in the source image, followed by employing an NMT model for translation. Finally, the text region in the source image is removed and rendered with translated target text.
The drawbacks of the cascade model mainly lie in:
(1) The cascade OCR-NMT procedure has the risk of error propagation and negatively affected translation quality.
(2) The removing and rendering process damages the integrity of background in the image, resulting in the suboptimal visual quality of the output image.

To address the issues in the existing cascade model, recent research has focused on the end-to-end IIMT model \cite{mansimov-etal-2020-e2eiimt, tian-etal-2023-pixelseq2seq, lan-etal-2024-translatotronv}.
As shown in Figure \ref{fig:example}, previous research mainly focuses on images with simple backgrounds, single font style, fixed text position, and bilingual translation.

However, real-world images may contain complex backgrounds, various font styles, diverse text positions, and a wide range of translation directions, which leads to certain limitations in previous research.
To overcome these limitations, we explore Practical In-Image Multilingual Machine Translation (IIMMT), which emphasizes two key aspects: real-world captured images and multilingual translation. 
We annotate a dataset PRIM to convince the lack of publicly available real-world data, which contains real-world captured source images with one-line text and manually annotated target images. The images of PRIM include real-world backgrounds, various font styles, diverse text positions, and supporting $5$ translation directions.

To tackle the challenge of practical conditions in the PRIM dataset, we design an end-to-end model \textbf{Vis}ual\textbf{Trans}lator (VisTrans), which handles the visual text and background information in the image separately, with a two-stage training and multi-task learning strategy.
The separate processing of visual text and background ensures that the model retains the multilingual translation capability while maintaining the integrity of the background, thereby helps mitigate error propagation and improves visual quality.

The main contributions of this paper are as follows:
\begin{itemize}
    \item In order to closely resembles practical conditions, we explore Practical In-Image Multilingual Machine Translation (IIMMT). The challenge of the task lies in the lack of publicly available dataset, and the model needs to handle complex real-world images with multilingual translation directions.
    \item To mitigate the lack of real-world dataset, we present the first annotated dataset PRIM, which contains real-world captured images with multilingual translation directions.
    \item We propose a novel model VisTrans, the first end-to-end model designed for practical conditions. By handling the visual text and background information of the images separately, our model is able to generate multilingual target images with integrity backgrounds.
\end{itemize}

\section{Related Work}
\begin{figure*}[t]
    \centering
    \includegraphics[width=0.95\linewidth]{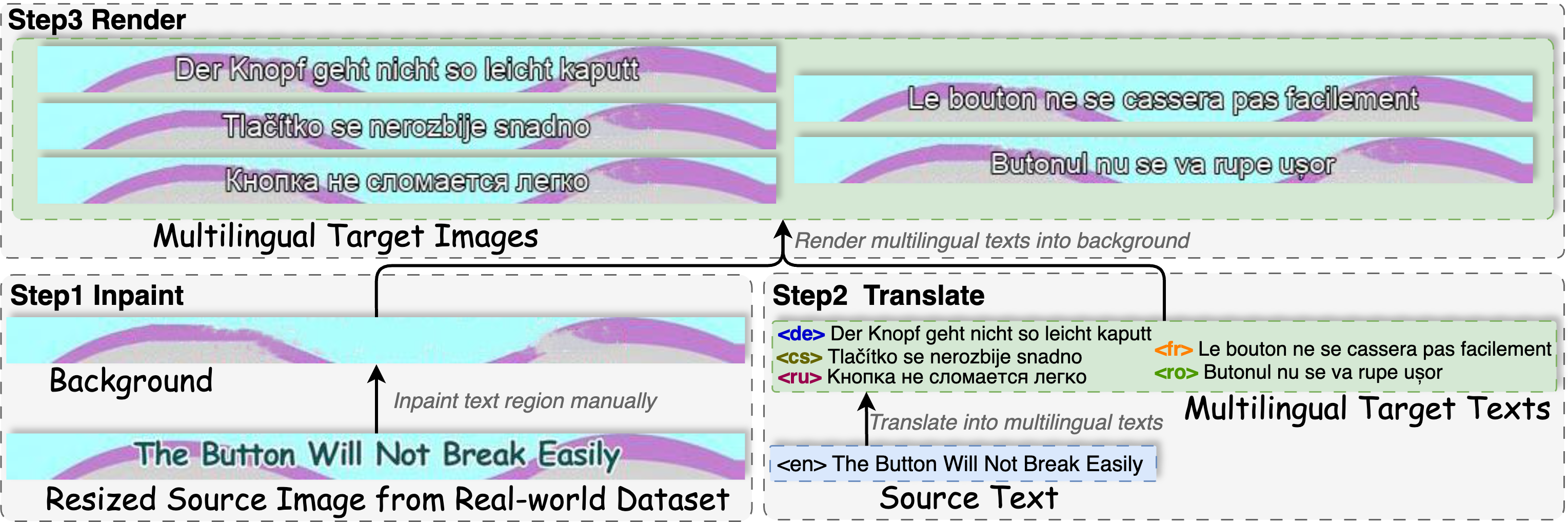}
    \caption{Annotation procedure of PRIM. The first step is inpainting the text region of the source image manually, obtaining the background. The second step is translating the source text into multilingual target texts. The third step is rendering the multilingual target texts into the background to get multilingual target images.}
    \label{fig:dataset}
\end{figure*}

\paragraph{In-Image Machine Translation.}
The end-to-end IIMT research mainly focuses on simple scenarios with synthetic datasets. 
\citet{mansimov-etal-2020-e2eiimt} and \citet{tian-etal-2023-pixelseq2seq} conduct on a dataset containing one-line black texts within a white background.
\citet{lan-etal-2024-translatotronv} further extend the dataset, including multiple lines of black text with random rotating in a solid-colored background.

Another type of research in IIMT does not focus on end-to-end models but instead utilizes existing pre-trained models to construct a more effective cascade model.
\citet{qian-etal-2024-anytrans} propose AnyTrans, an advanced pipeline that applies Qwen \cite{bai-etal-2023-qwen} instead of the NMT model, AnyText \cite{tuo-etal-2023-anytext} alternating Removing and Rendering procedures.
Although AnyTrans utilizes large pre-trained models, it is still constrained by the cascade process, which poses a risk of error propagation. It is also limited by the capability of the text editing model AnyText, making it hard to generate target images containing lengthy text.

\paragraph{Text-Image Translation.} Text-Image Translation (TIT) aims to translate the text in the image into the target text.
Research on TIT can be categorized into two main types: translating sentences within images \cite{lan-etal-2023-mctit, zhu-etal-2023-peit, ma-etal-2024-e2etit, li-etal-2024-mit10m}, and translating paragraph texts in images with layout information \cite{liang-etal-2024-documentit, zhang-etal-2025-layoutdit, liang-etal-2025-m4doc, liang-etal-2025-ssr, zhang-etal-2025-qrdit}.

\paragraph{Two-pass Model.}
The two-pass model is initially used for end-to-end Speech-to-Speech Translation (S2ST) task \cite{jia-etal-2022-translatotron2, inaguma-etal-2023-unity}, that the model firstly generate the target text with the source speech (1-pass), and the hidden representation of the target text is then used to generate the target spectrogram or the discrete code (2-pass).
\citet{fang-etal-2024-ComSpeech} investigate the vocabulary mismatch issue between the two decoders in the two-pass model, which makes it challenging to utilize existing Speech-to-Text Translation (S2TT) and Text-to-Speech (TTS) models, and propose a CTC \cite{graves-etal-2006-ctc} based vocabulary adaptor, and \citet{yao-etal-2024-deterministic} explore translation robustness under different vocabulary sizes.

\section{Data Construction}
\label{sec:testset}

\paragraph{PRIM.}
To address the lack of publicly available real-world IIMMT benchmark, we annotate a test set containing real-world source images with one-line texts and annotated target images, namely \textbf{Pr}actical In-\textbf{I}mage \textbf{M}ultilingual Machine Translation (PRIM).
The construction procedure is shown in Figure \ref{fig:dataset}.
To better align with real-world scenarios, PRIM follows the design paradigm commonly adopted in S2ST \cite{fang-etal-2023-daspeech, fang-etal-2024-ComSpeech, zhang-etal-2024-streamspeech}, such as CVSS \cite{jia-etal-2022-cvss}, which uses real-world human speech as source input and synthetic target speech. 

Recognizing the importance of realism on the source side, we adopt the same approach in PRIM, using real-world source images with annotated target images.
We take images collected from the real world by \citet{ma-etal-2024-e2etit} and \citet{li-etal-2024-mit10m} as source images, where both datasets are originally designed for TIT task.
\citet{ma-etal-2024-e2etit} capture images from video subtitles, and the English texts in the images are represented with different fonts, sizes, and positions.
\citet{li-etal-2024-mit10m} crawl textual images from websites and most of which are e-commerce platform advertising boards. 
Unlike the video subtitle scenarios, where the source text may be accessible through metadata, advertising boards typically do not provide such textual information. This scenario further emphasizes the necessity of the IIMT task, which aims to directly translate the input image into a target-language image without relying on the availability of source texts.

We crop the text regions of images in the above datasets with a size of Height $\times$ Width = $32 \times 512$ from the dataset and primarily select images with non-solid color backgrounds.
After source images are fully collected, the text areas are inpainting manually by the raster graphics editor software, obtaining the corresponding background of each source image.
Although PRIM only contains one-line text images captured from real world and does not fully reflect the real-world conditions, collecting and annotating such data remains highly challenging \cite{lan-etal-2024-translatotronv}. Compared with existing public datasets, PRIM offers a closer approximation to real-world conditions.

\begin{figure*}[t]
    \centering
    \includegraphics[width=0.95\linewidth]{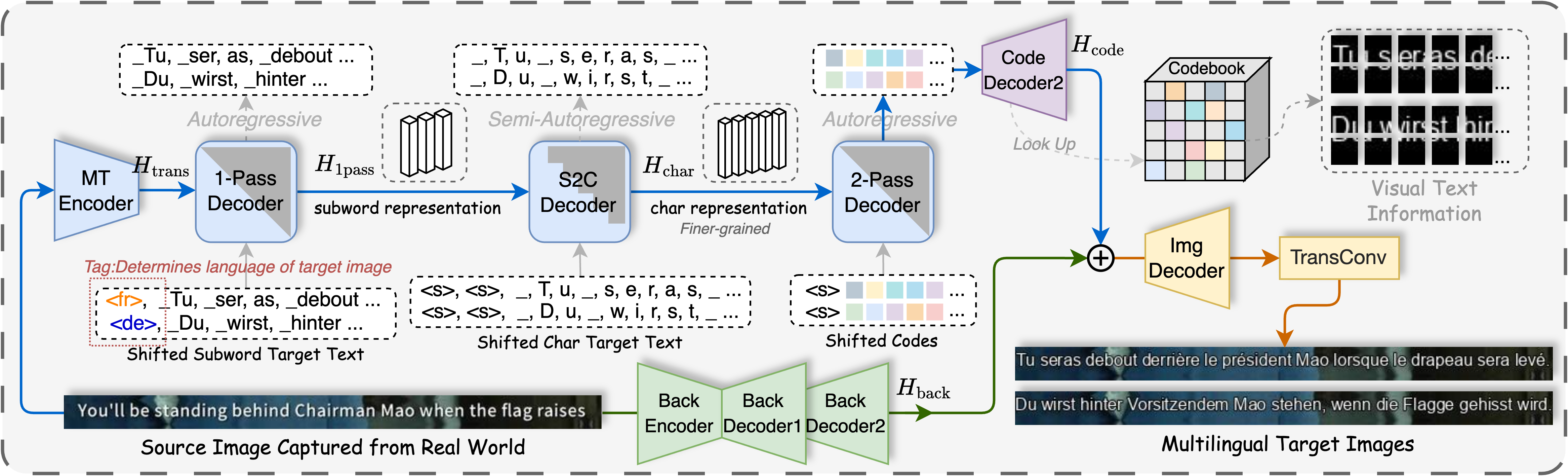}
    \caption{Architecture of VisTrans. The trapezoid represents Vision Transformer (ViT) \cite{dosovitskiy-etal-2021-vit}, while the rectangle represents Transformer Decoder \cite{vaswani-etal-2017-transformer}. More specifically, the rectangle with an upper triangular mask indicates an autoregressive decoder, whereas a staircase-like mask represents a semi-autoregressive decoder.}
    \label{fig:model}
\end{figure*}

We adopt a multilingual translation setting, focusing on one-to-many translation, which includes the following $5$ translation directions, English-Russian (En-Ru), English-French (En-Fr), English-Romanian (En-Ro), English-German (En-De), and English-Czech (En-Cs). 
The source texts are translated to multilingual target texts by GPT-4 \footnote{gpt-4-turbo with prompt \textit{``Translate the following sentence from English to \{target\_language\}: \{source\_text\}''}} and Google Translate, which are two commonly used methods to annotate translation texts \cite{li-etal-2024-mit10m, liang-etal-2024-documentit}.
The multilingual translated texts are then rendered into the background, to build the target images. 
By manually annotating target images, automatic evaluation metrics such as FID can be used to assess the visual quality by comparing the target images with the generated images \cite{tuo-etal-2023-anytext}. 

We perform inspections on PRIM, including the translation quality and integrity of image, and more details and evaluation results of PRIM are introduced in Appendix \ref{sec:appendix-testset}.

\paragraph{Training Set.}
Collecting and annotating large-scale IIMMT datasets from reality for training models is challenging, as the IIMMT dataset requires both source images and target images with parallel texts. 
Such a challenge also appears in the training of early OCR models, and one solution is to synthesize a large number of images using various fonts and backgrounds to simulate real-world scenarios \cite{jaderberg-etal-2014-syntheticocr}. 
Following the above method, we use TRDG toolkit \footnote{\url{https://github.com/Belval/TextRecognitionDataGenerator}} to render source texts with various styles, and the target images are rendered target texts using Arial font with different font sizes by PIL library.
The texts are sourced from the MTed dataset \cite{duh-2018-mted}, and the backgrounds of the source and target images are extracted from frames of the corresponding video based on the timestamps of the text.
Since image generation tasks typically use fixed input and output dimensions \cite{Esser-etal-2021-vqgan, Rombach-etal-2022-ldm, zhang-etal-2023-brushtext}, we extract the bottom part of the aforementioned frames with a size of Height $\times$ Width = $32 \times 512$.

The generated images are filtered to ensure that the source and target texts are fully rendered into the images. More details of the training set are introduced in Appendix \ref{sec:appendix-trainingset}.

\paragraph{Explanation of images with single-line texts.}
The images in our dataset contain single-line text with various font styles, sizes and positions, since the single-line text represents a fundamental and frequently encountered case in practical applications.
Prior research on TIT, such as PEIT \cite{zhu-etal-2023-peit} and METIMT \cite{ma-etal-2024-e2etit}, typically uses single-line source images as input. In particular, the constructed ECOIT dataset contains a large collection of single-line text images captured from an e-commerce platform, indicating that such images are present at scale in real-world scenarios.

\section{Method}
\label{sec:model}
We design the VisTrans model which is shown in Figure \ref{fig:model}. We first introduce the overall architecture and inference process of the model, where the output target image is generated given an input source image. Then, we describe the training methodology of the model.

\begin{figure*}[t]
    \centering
    \includegraphics[width=0.95\linewidth]{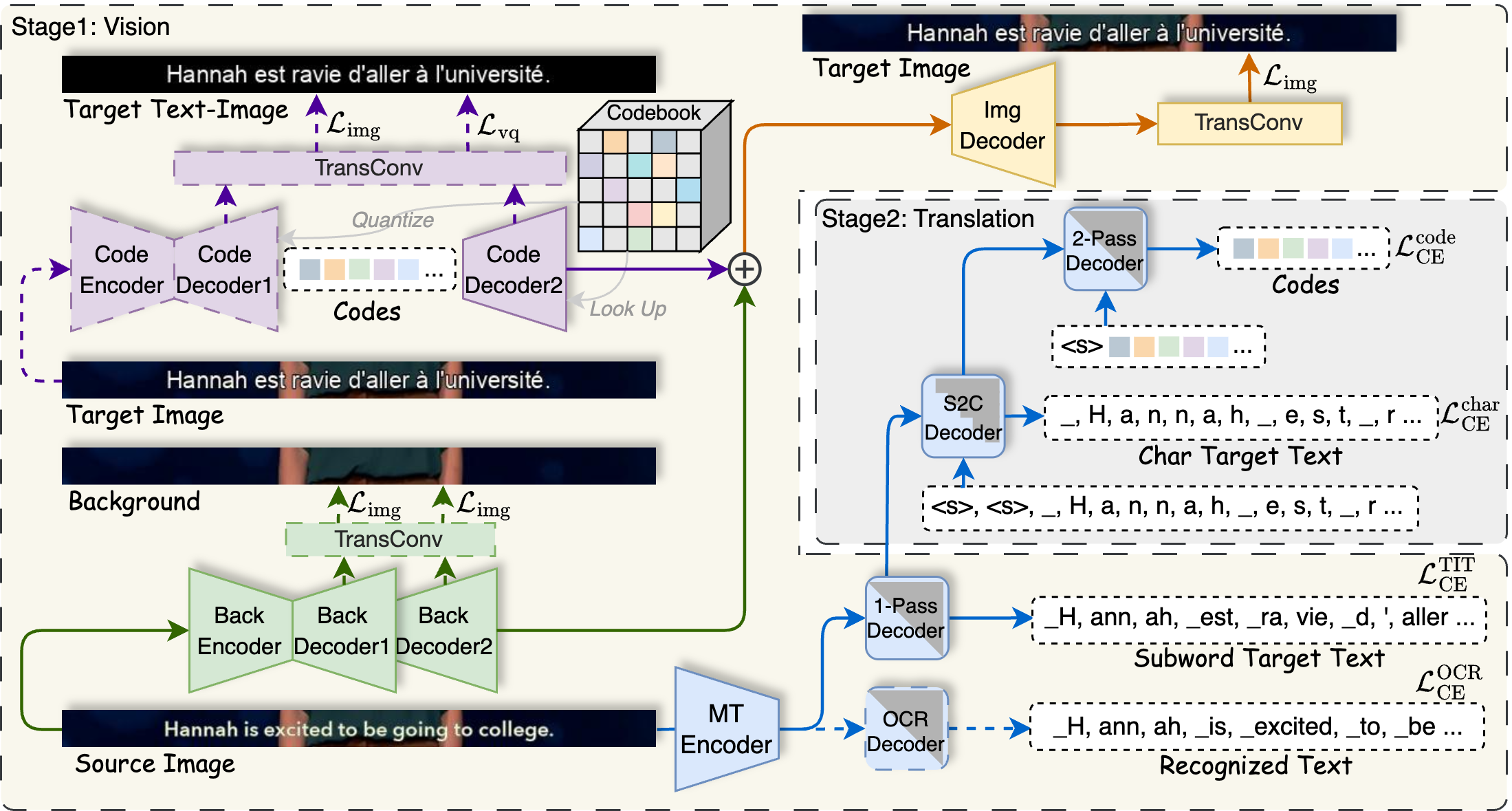}
    \caption{The two stages training of VisTrans. The modules and arrows with dashed lines represent auxiliary modules and tasks introduced during training, which will not be used during model inference.}
    \label{fig:train}
\end{figure*}

\subsection{Architecture}

The source image $I_\text{src} \in \mathbb{R} ^{H \times W \times C} $ is used as input for two sub-modules.
Firstly, it is encoded by a set of Vision Transformer (ViT) \cite{dosovitskiy-etal-2021-vit}, BackEncoder, BackDecoder1, and BackDecoder2, obtaining the output representation $H_\text{back} \in \mathbb{R} ^ {\frac{HW}{P^2} \times D}$, where $P$ and $D$ are the patch size and dimension of the ViT.
Secondly, it is encoded into representation $H_\text{trans} \in \mathbb{R} ^ {\frac{HW}{P^2} \times D}$ by another ViT, MT Encoder.

The 1-Pass Decoder is used to generate the target translation text, and its hidden representation $H_\text{1pass} \in \mathbb{R} ^ {L_s \times D} $ where $L_s$ and $D$ are the lengths of subword target text and dimension of the 1-Pass Decoder, serving as input for the subsequent module.
Specifically, the 1-Pass Decoder takes the embeddings of the shifted target text, which is prefixed with a language tag (e.g., ``<de>'' for German, ``<fr>'' for French), along with $H_\text{trans}$ as input, and then autoregressively generates the target language text and the hidden representation $H_\text{1pass}$.

The hidden representation $H_\text{1pass}$ is used as input of a Subword-to-Char (S2C) Decoder, which aims to transform the subword-level representation $H_\text{1pass}$ into the char-level representation $H_\text{char} \in \mathbb{R} ^ {L_c \times D}$ where $L_c$ is the length of char target text.
The S2C Decoder is implemented with a Semi-Autoregressive (SAT) decoder \cite{wang-wtal-2018-semiar}, which generates a group of $K$ tokens at each step, rather than producing a single token like an autoregressive (AT) decoder. Compared with the AT decoder, the SAT decoder has a certain performance degradation while achieving higher decoding efficiency.
The architectural difference between SAT and AT decoders lies in the design of the attention mask. AT typically employs a strict causal mask which visually appears as an upper triangle, whereas SAT utilizes a relaxed causal mask which visually appears as a staircase. 

The char-level representation $H_\text{char}$ is used to generate the code sequence with the 2-Pass Decoder autoregressively, and each code in the sequence is looked up with a codebook, converted to a vector sequence $H_\text{code}$, which is added with the output of BackDecoder2 $H_\text{back}$, to generate the target image with ImgDecoder and Transposed Convolution.

The reason for using the S2C Decoder to convert subword-level representation into char-level representation is to align with the representations in the codebook.
The designed codebook stores the visual text information of the target image, where each code corresponds to a small patch (e.g., $16 \times 16$) of the image.
Therefore, a single code is usually insufficient to fully capture the visual characteristics of the region corresponding to a subword.
In other words, the granularity of subword representation is too large compared to code representation, which requires transforming the subword into a finer-grained representation, to reduce the granularity gap between it and the code representation.

\subsection{Training}

We employ a multi-task learning strategy along with auxiliary modules to train our model, and the training process includes two stages, as illustrated in Figure \ref{fig:train}.
The solid-lined modules are required during inference, while the dashed-lined modules are auxiliary components used only for training.
The auxiliary modules aim to generate background, text-image (images containing visual texts with an empty background), and recognized text, which are not required during inference.
The following is an introduction to the three loss functions used in the training process.

The image reconstruction loss with perceptual loss \cite{zhang-etal-2018-perceptual} is used to train the image generation task, described as:
\begin{equation}
\label{eq:recloss}
    \mathcal{L}_\text{img}(y, \hat{y}) = ||y-\hat{y}||^2 +\lambda_\text{p}\mathcal{L}_\text{Perceptual}(y, \hat{y}),
\end{equation}
where $y$ is the generated image of the model and $\hat{y}$ is the ground truth image. $\lambda_\text{p}$ is loss weight, and it is set to $0.1$ in our experiments.

The vector quantization (vq) loss is used to train the codebook, formally as: 
\begin{equation}
\label{eq:vqcloss}
\begin{split}
    \mathcal{L}_\text{vq}(y, \hat{y}) = ||y-\hat{y}||^2 &+ \lambda_\text{p}\mathcal{L}_\text{Perceptual}(y, \hat{y}) \\ &+ ||\text{sg}[z_q] - E(x)||_2^2,
\end{split}
\end{equation}
where $y$ is the generated image of the model and $\hat{y}$ is the ground truth image.
$||\text{sg}[z_q] - E(x)||_2^2$ is the commitment loss with stop-gradient operation $\text{sg[·]}$, and $z_q$ is the vector obtained by quantization using the codebook, while $E(x)$ is the encoded feature of image $x$, which serves as the input to the quantization layer.
$\lambda_\text{p}$ is loss weight, and it is set to $0.1$ in our experiments.

The cross-entropy loss $\mathcal{L}_\text{CE}$ is used to train the sequence generation task, such as the generation of the target text and code sequence, and the cross entropy loss is applied label smoothing with $0.1$.

\paragraph{Stage 1: Vision.}  

In this stage, the model is trained in two parallel branches simultaneously.
The first branch is primarily designed to learn the visual text information in the target image.
The target image is used as input of the CodeEncoder and CodeDecoder1, obtaining the representation $E(x)$, which serves as the input of a codebook.

Specifically, the codebook $q$ contains $V$ learnable vectors $\{e_1, e_2, \ldots, e_V\}$, and each encoded feature $x_i$ is quantized by the nearest vector in $q$, obtaining $z_i$. Formally as:
\begin{equation}
    z_i = q(E(x_i)) = \argmin_{e_k \in q} || E(x_i) - e_k ||_2.
\end{equation}
With the quantization of the codebook, the target image is converted into a code sequence and the corresponding vector sequence. The vector sequence is further fed into the CodeDecoder2 to obtain the representation $H_\text{code}$.

To align the learnable vectors in the codebook with the visual text in the image, 
transposed convolutions are applied to generate output images $I_\text{code1}$ and $I_\text{code2}$ based on the hidden representations of CodeDecoder1 and CodeDecoder2, which are used to train the reconstruction of the target text-image $I_\text{tgt-text}$.
The loss function of the first branch is described as follows:
\begin{equation}
    \mathcal{L}_\text{b1} = \mathcal{L}_\text{img}(I_\text{code1}, I_\text{tgt-text}) + \mathcal{L}_\text{vq}(I_\text{code2}, I_\text{tgt-text}).
\end{equation}

The second branch is mainly used to learn the background information in the source image.
The hidden representation of the background $H_\text{back}$ is obtained by encoding the source image using a set of BackEncoder, BackDecoder1, and BackDecoder2. 
Similar to the first branch, the hidden representations of BackDecoder1 and BackDecoder2 are used to generate the output images $I_\text{back1}$ and $I_\text{back2}$ with transposed convolutions, aiming to reconstruct the background $I_\text{back}$.

In addition, the second branch also learns the information required for the training in the subsequent translation stage, an MT Encoder and a 1-Pass Decoder are used to generate the multilingual target texts (TIT task), with the OCR Decoder to recognize the texts in source images for auxiliary. The two tasks are trained with cross-entropy loss, denoted as $\mathcal{L}_\text{CE}^\text{TIT}$ and $\mathcal{L}_\text{CE}^\text{OCR}$.
The loss function of the second branch is described as follows:
\begin{equation}
\begin{split}
    \mathcal{L}_\text{b2} = \mathcal{L}_\text{img}(I_\text{back1}, I_\text{back}) &+ \mathcal{L}_\text{img}(I_\text{back2}, I_\text{back}) \\ &+ \mathcal{L}_\text{CE}^\text{TIT} + \mathcal{L}_\text{CE}^\text{OCR}.
\end{split}
\end{equation}

The final output representations of two branches, $H_\text{code}$ and $H_\text{back}$ are added, to generate the image $I_\text{img}$ by the ImgDecoder and Transposed Convolution, aiming to reconstruct the target image $I_\text{tgt}$.
The total loss function of stage 1 is the sum of all the loss functions mentioned above, expressed as:
\begin{equation}
    \mathcal{L}_\text{stage1} = \mathcal{L}_\text{b1} + \mathcal{L}_\text{b2} + \mathcal{L}_\text{img}(I_\text{img}, I_\text{tgt}).
\end{equation}

\begin{table*}[t]
    \centering
    \setlength{\tabcolsep}{5pt}
    \begin{tabular}{ccccccc|c}
    \Xhline{1.5pt}
        \multirow{2}{*}{\textbf{Systems}} & \multicolumn{6}{c}{\textbf{BLEU} $\uparrow$ / \textbf{COMET} $\uparrow$} & \textbf{FID} $\downarrow$ \\
        \cline{2-8}
         & \textbf{En-De} & \textbf{En-Fr} & \textbf{En-Cs} & \textbf{En-Ru} & \textbf{En-Ro} & \textbf{Avg.} & \textbf{Avg.} \\
    \hline
        {\small Golden} & {\small 74.2 / 89.6} & {\small 73.7 / 86.9} & {\small 73.3 / 89.7} & {\small 47.1 / 84.4} & {\small 65.1 / 75.3} & {\small 66.7 / 85.2} & {\small 0.00} \\
    \hline
    \multicolumn{8}{c}{\small\textit{pre-trained cascade models}} \\
    \hline
        {\small EasyOCR-NLLB-Render} & {\small 24.7 / 58.4} & {\small 27.0 / 60.4} & {\small 24.0 / 67.8} & {\small 14.1 / 60.8} & {\small 25.2 / 66.0} & {\small 23.0 / 62.7} & {\small 100.2}  \\
        {\small QwenVL-Render} & {\small 19.8 / 56.7} & {\small 23.3 / 56.9} & {\small 14.5 / 61.2} & {\small 11.3 / 56.8} & {\small 16.7 / 60.4} & {\small 17.1 / 58.4} & {\small 102.2} \\
        {\small AnyTrans} & {\small 0.1 / 29.8 } & {\small 0.1 / 30.6 } & {\small 0.0 / 30.9 } & {\small 0.1 / 32.4 } & {\small 0.0 / 31.1} & {\small 0.1 / 31.0} & {\small 204.1}\\
    \hline 
    \multicolumn{8}{c}{\small\textit{cascade models}} \\
    \hline
        {\small PARSeq-mTransformer-Render} & {\small 9.5 / 41.7} & {\small 13.8 / 46.9} & {\small \underline{7.7} / \underline{43.9}} & {\small \underline{5.5} / \underline{48.1}} & {\small 12.8 / \textbf{53.5}} & {\small 9.9 / 46.8} & {\small 103.8} \\
        {\small PEIT-Render} & {\small \underline{10.4} / \textbf{45.1}} & {\small \underline{14.0} / \underline{48.1}} & {\small \textbf{7.9} / \textbf{46.2}} & {\small 5.3 / 47.7} &{\small \textbf{14.2} / \underline{52.9}} & {\small \underline{10.4} / \textbf{48.0}} &  {\small 101.4} \\
    \hline
    \multicolumn{8}{c}{\small\textit{end-to-end models}} \\
    \hline
        {\small TranslatotronV} & {\small 1.7 / 34.3 } & {\small 1.9 / 30.2 } & {\small 1.1 / 30.5 } & {\small 0.9 / 32.0 } & {\small 1.3 / 33.9 } & {\small 1.4 / 32.2 } & {\small \underline{69.1}} \\
        {\small VisTrans (ours)} & {\small \textbf{12.6} / \underline{44.4}} & {\small \textbf{17.0} / \textbf{49.4}} & {\small 5.9 / 41.8 } & {\small \textbf{7.2} / \textbf{49.4}} & {\small \underline{13.9} / 50.2} & {\small \textbf{11.3} / \underline{47.0}} & {\small \textbf{28.8}} \\
    \Xhline{1.5pt}
    \end{tabular}
    \caption{Experimental results of different systems. Metrics include translation quality (BLEU, COMET) and visual effect (FID), and ``Avg.'' represents the average across all translation directions. $\uparrow$ or $\downarrow$ indicates higher or lower values are better. The \textbf{best} and \underline{second-best} performance are in bold and underline, respectively.}
    \label{tab:mainresult}
\end{table*}

\paragraph{Stage 2: Translation.}
The training in stage 1 is divided into two branches, which obtain the visual text information of the target image by the quantization of the codebook, and the background information of the source image. The final output image can be generated with the addition of the two types of information.
The background information can be obtained directly from the source image, but the code sequence corresponding to the visual text information cannot be directly obtained.

Therefore, the training objective of stage 2 is to generate the code sequence for the target image based on the representations from the pre-trained MT Encoder and 1-Pass Decoder in stage 1.
Specifically, the hidden representation of the 1-Pass Decoder and the embedding of the shifted char target text is inputted into the S2C Decoder, transforming the subword-level representation of target text into the char-level representation.
The char-level representation, along with the embedding of the shifted code sequence, is further used in the 2-Pass Decoder to generate the code sequence.
Both tasks are trained using cross-entropy loss, denoted as $\mathcal{L}_\text{CE}^\text{char}$, $\mathcal{L}_\text{CE}^\text{code}$. 
The complete loss function of stage 2 is the sum of them, expressed as:
\begin{equation}
    \mathcal{L}_\text{stage2} = \mathcal{L}_\text{CE}^\text{char} + \mathcal{L}_\text{CE}^\text{code}.
\end{equation}

\section{Experiments}

\subsection{Metrics}
The evaluation of IIMT requires to recognize the texts in the output images \cite{tian-etal-2023-pixelseq2seq, tian-etal-2025-debackx,lan-etal-2024-translatotronv, qian-etal-2024-anytrans}, and we use EasyOCR \footnote{\url{https://github.com/JaidedAI/EasyOCR}}, a widely used OCR toolkit that supports multilingual text recognition, to recognize the generated images of each system.
Based on the OCR recognition results and the reference texts, we calculate BLEU \cite{papineni-etal-2002-bleu} and COMET \cite{rei-etal-2020-comet} to assess the translation quality.
The BLEU is calculated with \texttt{SacreBLEU} \footnote{\url{https://github.com/mjpost/sacrebleu}}, and COMET is calculated with \texttt{Unbabel-COMET} \footnote{\url{https://github.com/Unbabel/COMET}} by \texttt{wmt22-comet-da} model.

To evaluate the visual effect of the output images automatically and objectively, we calculate the Fréchet Inception Distance (FID) between the generated images and the reference images.
The FID correlates well with human judgment of visual quality, and it is a widely used metric in image generation tasks \cite{Esser-etal-2021-vqgan, Rombach-etal-2022-ldm, Peebles-etal-2023-dit, tuo-etal-2023-anytext}. 
We employ \texttt{pytorch-fid} \footnote{\url{https://github.com/mseitzer/pytorch-fid}} to calculate the FID.

\begin{table*}[htbp]
    \centering
    \setlength{\tabcolsep}{5pt}
    \begin{tabular}{ccc}
    \Xhline{1.5pt}
        \textbf{Input Images} & \textbf{Systems} & \textbf{Output Images} \\
    \hline
    \addlinespace[2pt]
        \multirow{2}{*}{\includegraphics[width=0.42\textwidth]{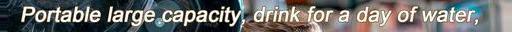}} & {\small Cascade} & \includegraphics[width=0.42\textwidth]{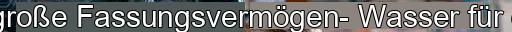} \\
         & {\small VisTrans} & \includegraphics[width=0.42\textwidth]{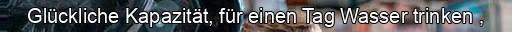}  \\
    \hline
    \addlinespace[2pt]
        \multirow{2}{*}{\includegraphics[width=0.42\textwidth]{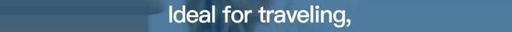}} & {\small Cascade} & \includegraphics[width=0.42\textwidth]{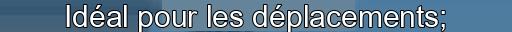}  \\
         & {\small VisTrans} & \includegraphics[width=0.42\textwidth]{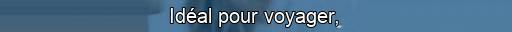}  \\
    \Xhline{1.5pt}
    \end{tabular}
    \caption{Comparison of cascade model (EasyOCR-NLLB-Render) and our end-to-end VisTrans. The issues of the cascade system output images lie in that the background of the images is damaged, negatively affecting the visual quality; and the texts are not fully rendered, decreasing the translation quality.}
    \label{tab:cases}
\end{table*}

\subsection{Experimental Settings}
We use PRIM mentioned in Section \ref{sec:testset} to evaluate the systems, including pre-trained cascade models, cascade models, and end-to-end models. 
The following is a brief introduction to each system.

\paragraph{Golden.} 
Since evaluating IIMMT models requires recognizing the texts in the output images with the OCR model, which could introduce errors, negatively impacting the evaluation of translation quality. We use the same OCR model to evaluate the golden reference target images in the test set, and these results represent the theoretical upper bound for all system.

\paragraph{EasyOCR-NLLB-Render.}  
A cascade system includes pre-trained EasyOCR, NLLB-3.3B \cite{nllb-2022}, and text render.
The text render first requires removing the texts in the source images, and we replace each text area detected by EasyOCR with the mean color of the region.
Then the translated texts are rendered in Arial font.
Unless otherwise specified, this render method is also used in other cascade models, which ensures the integrity of the images as much as possible. 

\paragraph{QwenVL-Render.}
A cascade system includes Qwen2.5VL-7B \cite{bai-etal-2025-qwen25vl} with prompt \textit{``Translate the text in the image to \{target\_language\}, and only output the translated text''}, and text render.

\paragraph{AnyTrans.} 
An advanced cascade model \cite{qian-etal-2024-anytrans} with pre-trained PPOCR, Qwen1.5-7B, and AnyText.
More explanation of AnyTrans is introduced in Appendix \ref{sec:anytrans}.

\paragraph{PARSeq-mTransformer-Render.} 
A cascade model contains SOTA OCR model PARSeq \cite{bautista-etal-2022-parseq}, a commonly used multilingual machine translation model mTransformer \cite{johnson-etal-2017-mtransformer}, and text render. 

\paragraph{PEIT-Render.}  
A cascade model contains the SOTA TIT model PEIT \cite{zhu-etal-2023-peit}, and text render. 

\paragraph{TranslatotronV.} 
An end-to-end IIMT model \cite{lan-etal-2024-translatotronv} with the architecture of ViT-VQGAN \cite{yu-etal-2022-vitvqgan} and multi-task learning.

\paragraph{VisTrans.} 
Our end-to-end IIMT model is introduced in Section \ref{sec:model}. Detailed implementation is introduced in Appendix \ref{sec:appendix-implementation-vistrans}.

\subsection{Main Results}
The experimental results are shown in Table \ref{tab:mainresult}.
\paragraph{Translation Quality.}
The EasyOCR-NLLB-Render achieves the best performance, which benefits from the strong translation performance of NLLB.
Except for pre-trained models, our VisTrans and PEIT-Render achieve better performance compared with other baselines.

\paragraph{Visual Effect.}
Although a more complex rendering method is applied for the cascade models (replacing text regions with average pixels instead of directly removing), the visual quality still remains poor.
The end-to-end models achieve better visual quality due to the incorporation of image generation modules. Compared to TranslatotronV, our VisTrans processes visual text and background information separately, leading to improved visual performance.

\section{Analysis}

\subsection{Ablation Study: Does S2C Decoder contribute to translation quality?}
The core idea of VisTrans is to handle the background and visual text separately, making all components except the S2C Decoder essential.
To investigate the different performance with the S2C Decoder, we conduct an ablation study, by replacing it with other type of decoder. The experimental results are shown in Table \ref{tab:ablation}.

\begin{table}[htbp]
    \centering
    \begin{tabular}{ccc}
    \Xhline{1.5pt}
    \textbf{S2C Decoder} & \textbf{Avg. BLEU $\uparrow$} & \textbf{Speedup $\uparrow$}\\
    \hline
    None & 5.27 & $1.00\times$ \\ 
    CTC & 6.06 & $0.96\times$\\
    AT & 11.87 &  $0.74\times$ \\    
    SAT (K=2) & 11.32 & $0.88\times$ \\
    SAT (K=4) & 8.97 & $0.92\times$ \\
    SAT (K=6) & 7.16 & $0.93\times$ \\
    \Xhline{1.5pt}
    \end{tabular}
    \caption{Average BLEU and Speedup on different S2C Decoder.}
    \label{tab:ablation}
\end{table}

Different types of S2C Decoder have an impact on translation quality. 
Specifically, by removing the S2C Decoder (None), the translation quality decreases significantly. The S2C Decoder with CTC is trained by upsampling the hidden representation from the 1-Pass Decoder, leading to an improvement compared to none S2C Decoder.
The best translation quality is achieved using AT for the S2C Decoder, but the autoregressive decoding results in slower inference speed.
As $K$ increases, SAT yields faster inference but lower translation quality. We adopt $K=2$ in the SAT S2C Decoder to balance quality and speed.

\subsection{Case Study: Why not render texts into images directly?}

Cascade models with text render have certain advantages: they leverage well-established text-based translation models, and the techniques for rendering texts into images are mature, ensuring clear and readable fonts.

However, as illustrated in Table \ref{tab:cases}, the cascade model (EasyOCR-NLLB-Render) causes noticeable damage to the output images, which is also reflected in the FID metric in Table \ref{tab:mainresult}.
Moreover, when render the translated text into the image, the text length is constrained by the image size, preventing complete rendering and leading to a decline in translation quality. 
Therefore, the error propagation in cascade models is not limited to the OCR-NMT process, and there still exists the issue of incomplete text rendering in the target image, which is another form of error propagation.
Since different font sizes are used for rendering the target images of the training set, our VisTrans can automatically adjust the font size in the output image, ensuring text completeness.
More output images of our VisTrans are shown in Appendix \ref{sec:moreoutputs}.

\subsection{Robustness Study: Does VisTrans fit for images containing multi-line texts?}

To evaluate the ability of the VisTrans model translating multi-line text images, we conduct experiments on the IIMT30k dataset \cite{tian-etal-2025-debackx}, which consists of synthetic images, but features complex backgrounds, diverse font styles, and a mix of single-line and multi-line texts (e.g., sentences split across two lines).

\begin{table}[htbp]
    \centering
    \begin{tabular}{ccccc}
    \Xhline{1.5pt}
    \multirow{2}{*}{\textbf{Systems}} & \multicolumn{2}{c}{\textbf{De-En}} & \multicolumn{2}{c}{\textbf{En-De}} \\
    & \textbf{Valid} & \textbf{Test} & \textbf{Valid} & \textbf{Test} \\
    \hline
    DebackX & 10.8 & 8.6 & 9.5 & 6.9 \\
    VisTrans (ours) & \textbf{14.7} & \textbf{12.3} & \textbf{16.5} & \textbf{12.2} \\
    \Xhline{1.5pt}
    \end{tabular}
    \caption{BLEU score on IIMT30k dataset.}
    \label{tab:iimt30k}
\end{table}

Experimental results in Table \ref{tab:iimt30k} demonstrate that VisTrans is capable of handling images containing multi-line text, indicating that the model can generalize to more complex text layouts when appropriately trained.

\section{Conclusion}
In this paper, we address the limitations of IIMT in real-world scenarios by exploring Practical In-Image Multilingual Machine Translation (IIMMT), and first annotate a dataset PRIM containing real-world images with multilingual translation directions.
To tackle the challenge of practical conditions in the PRIM dataset, we propose an end-to-end model VisTrans, which handles the visual text and background information separately. 
Experimental results show that our model retrains the multilingual translation capability while maintaining the integrity of the background, obtaining a better translation quality and visual effect compared to other models.

\section*{Limitations}
While we explore IIMMT by annotating PRIM dataset, and propose an end-to-end model VisTrans, this paper has certain limitations.

Our VisTrans is trained on large amount of training data, and is adopted a two-stage training with multi-task learning strategy, leading to the high computational resource costs and hardware requirements.
The images are quantized by a codebook, obtaining the code sequence. We only conduct experiments with the most basic codebook and decoder, lacking the investigation on the use of more advanced quantization techniques or decoders that better support long-sequence modeling.

\section*{Ethics Statement}
We manually annotate PRIM dataset containing real-world images, and the data has been carefully selected to avoid any form of offensive or biased
content. We take ethical considerations seriously and ensure that the data used in this study are conducted in a responsible and ethical manner.

\section*{Acknowledgment}
We thank all the anonymous reviewers for their insightful and valuable comments. This work is supported by the National Natural Science Foundation of China (Grant No. 62376027, 62406015) and Beijing Institute of Technology Science and Technology Innovation Plan (Grant No. 23CX13027).

\bibliography{custom}

\begin{thebibliography}{50}
\providecommand{\natexlab}[1]{#1}

\bibitem[{Bai et~al.(2023)Bai, Bai, Chu, Cui, Dang, Deng, Fan, Ge, Han, Huang, Hui, Ji, Li, Lin, Lin, Liu, Liu, Lu, Lu, Ma, Men, Ren, Ren, Tan, Tan, Tu, Wang, Wang, Wang, Wu, Xu, Xu, Yang, Yang, Yang, Yang, Yao, Yu, Yuan, Yuan, Zhang, Zhang, Zhang, Zhang, Zhou, Zhou, Zhou, and Zhu}]{bai-etal-2023-qwen}
Jinze Bai, Shuai Bai, Yunfei Chu, Zeyu Cui, Kai Dang, Xiaodong Deng, Yang Fan, Wenbin Ge, Yu~Han, Fei Huang, Binyuan Hui, Luo Ji, Mei Li, Junyang Lin, Runji Lin, Dayiheng Liu, Gao Liu, Chengqiang Lu, Keming Lu, Jianxin Ma, Rui Men, Xingzhang Ren, Xuancheng Ren, Chuanqi Tan, Sinan Tan, Jianhong Tu, Peng Wang, Shijie Wang, Wei Wang, Shengguang Wu, Benfeng Xu, Jin Xu, An~Yang, Hao Yang, Jian Yang, Shusheng Yang, Yang Yao, Bowen Yu, Hongyi Yuan, Zheng Yuan, Jianwei Zhang, Xingxuan Zhang, Yichang Zhang, Zhenru Zhang, Chang Zhou, Jingren Zhou, Xiaohuan Zhou, and Tianhang Zhu. 2023.
\newblock Qwen technical report.
\newblock \emph{arXiv preprint arXiv:2309.16609}.

\bibitem[{Bai et~al.(2025)Bai, Chen, Liu, Wang, Ge, Song, Dang, Wang, Wang, Tang, Zhong, Zhu, Yang, Li, Wan, Wang, Ding, Fu, Xu, Ye, Zhang, Xie, Cheng, Zhang, Yang, Xu, and Lin}]{bai-etal-2025-qwen25vl}
Shuai Bai, Keqin Chen, Xuejing Liu, Jialin Wang, Wenbin Ge, Sibo Song, Kai Dang, Peng Wang, Shijie Wang, Jun Tang, Humen Zhong, Yuanzhi Zhu, Mingkun Yang, Zhaohai Li, Jianqiang Wan, Pengfei Wang, Wei Ding, Zheren Fu, Yiheng Xu, Jiabo Ye, Xi~Zhang, Tianbao Xie, Zesen Cheng, Hang Zhang, Zhibo Yang, Haiyang Xu, and Junyang Lin. 2025.
\newblock \href {https://arxiv.org/abs/2502.13923} {Qwen2.5-vl technical report}.
\newblock \emph{Preprint}, arXiv:2502.13923.

\bibitem[{Bautista and Atienza(2022)}]{bautista-etal-2022-parseq}
Darwin Bautista and Rowel Atienza. 2022.
\newblock \href {https://doi.org/10.1007/978-3-031-19815-1_11} {Scene text recognition with permuted autoregressive sequence models}.
\newblock In \emph{European Conference on Computer Vision}, pages 178--196, Cham. Springer Nature Switzerland.

\bibitem[{Chen et~al.(2025)Chen, Song, Chen, Yang, Zhao, and Zhang}]{chen-etal-2025-clearerMMT}
Andong Chen, Yuchen Song, Kehai Chen, Muyun Yang, Tiejun Zhao, and Min Zhang. 2025.
\newblock \href {https://arxiv.org/abs/2412.12627} {Make imagination clearer! stable diffusion-based visual imagination for multimodal machine translation}.
\newblock \emph{Preprint}, arXiv:2412.12627.

\bibitem[{Dosovitskiy et~al.(2021)Dosovitskiy, Beyer, Kolesnikov, Weissenborn, Zhai, Unterthiner, Dehghani, Minderer, Heigold, Gelly, Uszkoreit, and Houlsby}]{dosovitskiy-etal-2021-vit}
Alexey Dosovitskiy, Lucas Beyer, Alexander Kolesnikov, Dirk Weissenborn, Xiaohua Zhai, Thomas Unterthiner, Mostafa Dehghani, Matthias Minderer, Georg Heigold, Sylvain Gelly, Jakob Uszkoreit, and Neil Houlsby. 2021.
\newblock \href {https://openreview.net/forum?id=YicbFdNTTy} {An image is worth 16x16 words: Transformers for image recognition at scale}.
\newblock In \emph{International Conference on Learning Representations}.

\bibitem[{Duh(2018)}]{duh-2018-mted}
Kevin Duh. 2018.
\newblock The multitarget ted talks task.
\newblock \url{http://www.cs.jhu.edu/~kevinduh/a/multitarget-tedtalks/}.

\bibitem[{Esser et~al.(2021)Esser, Rombach, and Ommer}]{Esser-etal-2021-vqgan}
Patrick Esser, Robin Rombach, and Bjorn Ommer. 2021.
\newblock Taming transformers for high-resolution image synthesis.
\newblock In \emph{Proceedings of the IEEE/CVF Conference on Computer Vision and Pattern Recognition (CVPR)}, pages 12873--12883.

\bibitem[{Fang and Feng(2022)}]{fang-etal-2022-phraseMMT}
Qingkai Fang and Yang Feng. 2022.
\newblock \href {https://doi.org/10.18653/v1/2022.acl-long.390} {Neural machine translation with phrase-level universal visual representations}.
\newblock In \emph{Proceedings of the 60th Annual Meeting of the Association for Computational Linguistics (Volume 1: Long Papers)}, pages 5687--5698, Dublin, Ireland. Association for Computational Linguistics.

\bibitem[{Fang et~al.(2024)Fang, Zhang, Ma, Zhang, and Feng}]{fang-etal-2024-ComSpeech}
Qingkai Fang, Shaolei Zhang, Zhengrui Ma, Min Zhang, and Yang Feng. 2024.
\newblock \href {https://doi.org/10.18653/v1/2024.acl-long.392} {Can we achieve high-quality direct speech-to-speech translation without parallel speech data?}
\newblock In \emph{Proceedings of the 62nd Annual Meeting of the Association for Computational Linguistics (Volume 1: Long Papers)}, pages 7264--7277, Bangkok, Thailand. Association for Computational Linguistics.

\bibitem[{Fang et~al.(2023)Fang, Zhou, and Feng}]{fang-etal-2023-daspeech}
Qingkai Fang, Yan Zhou, and Yang Feng. 2023.
\newblock \href {https://proceedings.neurips.cc/paper_files/paper/2023/file/e5b1c0d4866f72393c522c8a00eed4eb-Paper-Conference.pdf} {Daspeech: Directed acyclic transformer for fast and high-quality speech-to-speech translation}.
\newblock In \emph{Advances in Neural Information Processing Systems}, volume~36, pages 72604--72623. Curran Associates, Inc.

\bibitem[{Graves et~al.(2006)Graves, Fern\'{a}ndez, Gomez, and Schmidhuber}]{graves-etal-2006-ctc}
Alex Graves, Santiago Fern\'{a}ndez, Faustino Gomez, and J\"{u}rgen Schmidhuber. 2006.
\newblock \href {https://doi.org/10.1145/1143844.1143891} {Connectionist temporal classification: labelling unsegmented sequence data with recurrent neural networks}.
\newblock In \emph{Proceedings of the 23rd International Conference on Machine Learning}, ICML '06, page 369–376, New York, NY, USA. Association for Computing Machinery.

\bibitem[{Gugger et~al.(2022)Gugger, Debut, Wolf, Schmid, Mueller, Mangrulkar, Sun, and Bossan}]{accelerate}
Sylvain Gugger, Lysandre Debut, Thomas Wolf, Philipp Schmid, Zachary Mueller, Sourab Mangrulkar, Marc Sun, and Benjamin Bossan. 2022.
\newblock Accelerate: Training and inference at scale made simple, efficient and adaptable.
\newblock \url{https://github.com/huggingface/accelerate}.

\bibitem[{Inaguma et~al.(2023)Inaguma, Popuri, Kulikov, Chen, Wang, Chung, Tang, Lee, Watanabe, and Pino}]{inaguma-etal-2023-unity}
Hirofumi Inaguma, Sravya Popuri, Ilia Kulikov, Peng-Jen Chen, Changhan Wang, Yu-An Chung, Yun Tang, Ann Lee, Shinji Watanabe, and Juan Pino. 2023.
\newblock \href {https://doi.org/10.18653/v1/2023.acl-long.872} {{U}nit{Y}: Two-pass direct speech-to-speech translation with discrete units}.
\newblock In \emph{Proceedings of the 61st Annual Meeting of the Association for Computational Linguistics (Volume 1: Long Papers)}, pages 15655--15680, Toronto, Canada. Association for Computational Linguistics.

\bibitem[{Jaderberg et~al.(2014)Jaderberg, Simonyan, Vedaldi, and Zisserman}]{jaderberg-etal-2014-syntheticocr}
Max Jaderberg, Karen Simonyan, Andrea Vedaldi, and Andrew Zisserman. 2014.
\newblock \href {https://arxiv.org/abs/1406.2227} {Synthetic data and artificial neural networks for natural scene text recognition}.
\newblock \emph{Preprint}, arXiv:1406.2227.

\bibitem[{Jia et~al.(2022{\natexlab{a}})Jia, Ramanovich, Remez, and Pomerantz}]{jia-etal-2022-translatotron2}
Ye~Jia, Michelle~Tadmor Ramanovich, Tal Remez, and Roi Pomerantz. 2022{\natexlab{a}}.
\newblock \href {https://proceedings.mlr.press/v162/jia22b.html} {Translatotron 2: High-quality direct speech-to-speech translation with voice preservation}.
\newblock In \emph{Proceedings of the 39th International Conference on Machine Learning}, volume 162 of \emph{Proceedings of Machine Learning Research}, pages 10120--10134. PMLR.

\bibitem[{Jia et~al.(2022{\natexlab{b}})Jia, Tadmor~Ramanovich, Wang, and Zen}]{jia-etal-2022-cvss}
Ye~Jia, Michelle Tadmor~Ramanovich, Quan Wang, and Heiga Zen. 2022{\natexlab{b}}.
\newblock \href {https://aclanthology.org/2022.lrec-1.720/} {{CVSS} corpus and massively multilingual speech-to-speech translation}.
\newblock In \emph{Proceedings of the Thirteenth Language Resources and Evaluation Conference}, pages 6691--6703, Marseille, France. European Language Resources Association.

\bibitem[{Johnson et~al.(2017)Johnson, Schuster, Le, Krikun, Wu, Chen, Thorat, Vi{\'e}gas, Wattenberg, Corrado et~al.}]{johnson-etal-2017-mtransformer}
Melvin Johnson, Mike Schuster, Quoc~V Le, Maxim Krikun, Yonghui Wu, Zhifeng Chen, Nikhil Thorat, Fernanda Vi{\'e}gas, Martin Wattenberg, Greg Corrado, et~al. 2017.
\newblock Google’s multilingual neural machine translation system: Enabling zero-shot translation.
\newblock \emph{Transactions of the Association for Computational Linguistics}, 5:339--351.

\bibitem[{Lan et~al.(2024)Lan, Niu, Meng, Zhou, Zhang, and Su}]{lan-etal-2024-translatotronv}
Zhibin Lan, Liqiang Niu, Fandong Meng, Jie Zhou, Min Zhang, and Jinsong Su. 2024.
\newblock \href {https://doi.org/10.18653/v1/2024.findings-acl.325} {Translatotron-{V}(ison): An end-to-end model for in-image machine translation}.
\newblock In \emph{Findings of the Association for Computational Linguistics: ACL 2024}, pages 5472--5485, Bangkok, Thailand. Association for Computational Linguistics.

\bibitem[{Lan et~al.(2023)Lan, Yu, Li, Zhang, Luan, Wang, Huang, and Su}]{lan-etal-2023-mctit}
Zhibin Lan, Jiawei Yu, Xiang Li, Wen Zhang, Jian Luan, Bin Wang, Degen Huang, and Jinsong Su. 2023.
\newblock \href {https://doi.org/10.18653/v1/2023.acl-long.192} {Exploring better text image translation with multimodal codebook}.
\newblock In \emph{Proceedings of the 61st Annual Meeting of the Association for Computational Linguistics (Volume 1: Long Papers)}, pages 3479--3491, Toronto, Canada. Association for Computational Linguistics.

\bibitem[{Li et~al.(2025{\natexlab{a}})Li, Zhu, and Wen}]{li-etal-2024-mit10m}
Bo~Li, Shaolin Zhu, and Lijie Wen. 2025{\natexlab{a}}.
\newblock \href {https://aclanthology.org/2025.coling-main.346/} {{MIT}-10{M}: A large scale parallel corpus of multilingual image translation}.
\newblock In \emph{Proceedings of the 31st International Conference on Computational Linguistics}, pages 5154--5167, Abu Dhabi, UAE. Association for Computational Linguistics.

\bibitem[{Li et~al.(2025{\natexlab{b}})Li, Guo, Yao, Liu, and Wang}]{li-etal-2025-homebench}
Silin Li, Yuhang Guo, Jiashu Yao, Zeming Liu, and Haifeng Wang. 2025{\natexlab{b}}.
\newblock \href {https://doi.org/10.18653/v1/2025.acl-long.597} {$homebench$: Evaluating {LLM}s in smart homes with valid and invalid instructions across single and multiple devices}.
\newblock In \emph{Proceedings of the 63rd Annual Meeting of the Association for Computational Linguistics (Volume 1: Long Papers)}, pages 12230--12250, Vienna, Austria. Association for Computational Linguistics.

\bibitem[{Liang et~al.(2024)Liang, Zhang, Ma, Zhang, Zhao, Xiang, Zong, and Zhou}]{liang-etal-2024-documentit}
Yupu Liang, Yaping Zhang, Cong Ma, Zhiyang Zhang, Yang Zhao, Lu~Xiang, Chengqing Zong, and Yu~Zhou. 2024.
\newblock \href {https://doi.org/10.18653/v1/2024.naacl-long.392} {Document image machine translation with dynamic multi-pre-trained models assembling}.
\newblock In \emph{Proceedings of the 2024 Conference of the North American Chapter of the Association for Computational Linguistics: Human Language Technologies (Volume 1: Long Papers)}, pages 7084--7095, Mexico City, Mexico. Association for Computational Linguistics.

\bibitem[{Liang et~al.(2025{\natexlab{a}})Liang, Zhang, Zhang, Chen, Zhao, Xiang, Zong, and Zhou}]{liang-etal-2025-ssr}
Yupu Liang, Yaping Zhang, Zhiyang Zhang, Zhiyuan Chen, Yang Zhao, Lu~Xiang, Chengqing Zong, and Yu~Zhou. 2025{\natexlab{a}}.
\newblock \href {https://doi.org/10.18653/v1/2025.findings-acl.1213} {Improving {MLLM}{'}s document image machine translation via synchronously self-reviewing its {OCR} proficiency}.
\newblock In \emph{Findings of the Association for Computational Linguistics: ACL 2025}, pages 23659--23678, Vienna, Austria. Association for Computational Linguistics.

\bibitem[{Liang et~al.(2025{\natexlab{b}})Liang, Zhang, Zhang, Zhao, Xiang, Zong, and Zhou}]{liang-etal-2025-m4doc}
Yupu Liang, Yaping Zhang, Zhiyang Zhang, Yang Zhao, Lu~Xiang, Chengqing Zong, and Yu~Zhou. 2025{\natexlab{b}}.
\newblock \href {https://doi.org/10.18653/v1/2025.acl-long.606} {Single-to-mix modality alignment with multimodal large language model for document image machine translation}.
\newblock In \emph{Proceedings of the 63rd Annual Meeting of the Association for Computational Linguistics (Volume 1: Long Papers)}, pages 12391--12408, Vienna, Austria. Association for Computational Linguistics.

\bibitem[{Loshchilov and Hutter(2019)}]{loshchilov-etal-2018-adamw}
Ilya Loshchilov and Frank Hutter. 2019.
\newblock \href {https://openreview.net/forum?id=Bkg6RiCqY7} {Decoupled weight decay regularization}.
\newblock In \emph{International Conference on Learning Representations}.

\bibitem[{Lu et~al.(2025)Lu, Yu, Wang, Liu, Su, Liu, Guo, Liang, Wang, and Wang}]{lu-etal-2025-transbench}
Yuheng Lu, Qian Yu, Hongru Wang, Zeming Liu, Wei Su, Yanping Liu, Yuhang Guo, Maocheng Liang, Yunhong Wang, and Haifeng Wang. 2025.
\newblock \href {https://doi.org/10.18653/v1/2025.findings-acl.645} {{T}rans{B}ench: Breaking barriers for transferable graphical user interface agents in dynamic digital environments}.
\newblock In \emph{Findings of the Association for Computational Linguistics: ACL 2025}, pages 12464--12478, Vienna, Austria. Association for Computational Linguistics.

\bibitem[{Ma et~al.(2024)Ma, Han, Wu, Zhang, Zhao, Zhou, and Zong}]{ma-etal-2024-e2etit}
Cong Ma, Xu~Han, Linghui Wu, Yaping Zhang, Yang Zhao, Yu~Zhou, and Chengqing Zong. 2024.
\newblock \href {https://doi.org/10.1109/TASLP.2023.3324540} {Modal contrastive learning based end-to-end text image machine translation}.
\newblock \emph{IEEE/ACM Transactions on Audio, Speech, and Language Processing}, 32:2153--2165.

\bibitem[{Mansimov et~al.(2020)Mansimov, Stern, Chen, Firat, Uszkoreit, and Jain}]{mansimov-etal-2020-e2eiimt}
Elman Mansimov, Mitchell Stern, Mia Chen, Orhan Firat, Jakob Uszkoreit, and Puneet Jain. 2020.
\newblock \href {https://doi.org/10.18653/v1/2020.nlpbt-1.8} {Towards end-to-end in-image neural machine translation}.
\newblock In \emph{Proceedings of the First International Workshop on Natural Language Processing Beyond Text}, pages 70--74, Online. Association for Computational Linguistics.

\bibitem[{Niu et~al.(2024)Niu, Meng, and Zhou}]{niu-etal-2024-umtit}
Liqiang Niu, Fandong Meng, and Jie Zhou. 2024.
\newblock \href {https://aclanthology.org/2024.lrec-main.1474/} {{UMTIT}: Unifying recognition, translation, and generation for multimodal text image translation}.
\newblock In \emph{Proceedings of the 2024 Joint International Conference on Computational Linguistics, Language Resources and Evaluation (LREC-COLING 2024)}, pages 16953--16972, Torino, Italia. ELRA and ICCL.

\bibitem[{{NLLB Team} et~al.(2022){NLLB Team}, Costa-jussà, Cross, Çelebi, Elbayad, Heafield, Heffernan, Kalbassi, Lam, Licht, Maillard, Sun, Wang, Wenzek, Youngblood, Akula, Barrault, Gonzalez, Hansanti, Hoffman, Jarrett, Sadagopan, Rowe, Spruit, Tran, Andrews, Ayan, Bhosale, Edunov, Fan, Gao, Goswami, Guzmán, Koehn, Mourachko, Ropers, Saleem, Schwenk, and Wang}]{nllb-2022}
{NLLB Team}, Marta~R. Costa-jussà, James Cross, Onur Çelebi, Maha Elbayad, Kenneth Heafield, Kevin Heffernan, Elahe Kalbassi, Janice Lam, Daniel Licht, Jean Maillard, Anna Sun, Skyler Wang, Guillaume Wenzek, Al~Youngblood, Bapi Akula, Loic Barrault, Gabriel~Mejia Gonzalez, Prangthip Hansanti, John Hoffman, Semarley Jarrett, Kaushik~Ram Sadagopan, Dirk Rowe, Shannon Spruit, Chau Tran, Pierre Andrews, Necip~Fazil Ayan, Shruti Bhosale, Sergey Edunov, Angela Fan, Cynthia Gao, Vedanuj Goswami, Francisco Guzmán, Philipp Koehn, Alexandre Mourachko, Christophe Ropers, Safiyyah Saleem, Holger Schwenk, and Jeff Wang. 2022.
\newblock \href {https://arxiv.org/abs/2207.04672} {No language left behind: Scaling human-centered machine translation}.
\newblock \emph{Preprint}, arXiv:2207.04672.

\bibitem[{Papineni et~al.(2002)Papineni, Roukos, Ward, and Zhu}]{papineni-etal-2002-bleu}
Kishore Papineni, Salim Roukos, Todd Ward, and Wei-Jing Zhu. 2002.
\newblock \href {https://doi.org/10.3115/1073083.1073135} {{B}leu: a method for automatic evaluation of machine translation}.
\newblock In \emph{Proceedings of the 40th Annual Meeting of the Association for Computational Linguistics}, pages 311--318, Philadelphia, Pennsylvania, USA. Association for Computational Linguistics.

\bibitem[{Peebles and Xie(2023)}]{Peebles-etal-2023-dit}
William Peebles and Saining Xie. 2023.
\newblock Scalable diffusion models with transformers.
\newblock In \emph{Proceedings of the IEEE/CVF International Conference on Computer Vision (ICCV)}, pages 4195--4205.

\bibitem[{Qian et~al.(2024)Qian, Zhang, Yang, Fan, Ma, Wong, Sun, and Ji}]{qian-etal-2024-anytrans}
Zhipeng Qian, Pei Zhang, Baosong Yang, Kai Fan, Yiwei Ma, Derek~F. Wong, Xiaoshuai Sun, and Rongrong Ji. 2024.
\newblock \href {https://doi.org/10.18653/v1/2024.findings-emnlp.137} {{A}ny{T}rans: Translate {A}ny{T}ext in the image with large scale models}.
\newblock In \emph{Findings of the Association for Computational Linguistics: EMNLP 2024}, pages 2432--2444, Miami, Florida, USA. Association for Computational Linguistics.

\bibitem[{Rei et~al.(2020)Rei, Stewart, Farinha, and Lavie}]{rei-etal-2020-comet}
Ricardo Rei, Craig Stewart, Ana~C Farinha, and Alon Lavie. 2020.
\newblock \href {https://doi.org/10.18653/v1/2020.emnlp-main.213} {{COMET}: A neural framework for {MT} evaluation}.
\newblock In \emph{Proceedings of the 2020 Conference on Empirical Methods in Natural Language Processing (EMNLP)}, pages 2685--2702, Online. Association for Computational Linguistics.

\bibitem[{Rombach et~al.(2022)Rombach, Blattmann, Lorenz, Esser, and Ommer}]{Rombach-etal-2022-ldm}
Robin Rombach, Andreas Blattmann, Dominik Lorenz, Patrick Esser, and Bj\"orn Ommer. 2022.
\newblock High-resolution image synthesis with latent diffusion models.
\newblock In \emph{Proceedings of the IEEE/CVF Conference on Computer Vision and Pattern Recognition (CVPR)}, pages 10684--10695.

\bibitem[{Tian et~al.(2023)Tian, Li, Liu, Guo, and Wang}]{tian-etal-2023-pixelseq2seq}
Yanzhi Tian, Xiang Li, Zeming Liu, Yuhang Guo, and Bin Wang. 2023.
\newblock \href {https://doi.org/10.18653/v1/2023.findings-emnlp.1004} {In-image neural machine translation with segmented pixel sequence-to-sequence model}.
\newblock In \emph{Findings of the Association for Computational Linguistics: EMNLP 2023}, pages 15046--15057, Singapore. Association for Computational Linguistics.

\bibitem[{Tian et~al.(2025)Tian, Liu, Liu, and Guo}]{tian-etal-2025-debackx}
Yanzhi Tian, Zeming Liu, Zhengyang Liu, and Yuhang Guo. 2025.
\newblock \href {https://aclanthology.org/2025.findings-acl.6/} {Exploring in-image machine translation with real-world background}.
\newblock In \emph{Findings of the Association for Computational Linguistics: ACL 2025}, pages 124--137, Vienna, Austria. Association for Computational Linguistics.

\bibitem[{Tuo et~al.(2024)Tuo, Xiang, He, Geng, and Xie}]{tuo-etal-2023-anytext}
Yuxiang Tuo, Wangmeng Xiang, Jun-Yan He, Yifeng Geng, and Xuansong Xie. 2024.
\newblock \href {https://openreview.net/forum?id=ezBH9WE9s2} {Anytext: Multilingual visual text generation and editing}.
\newblock In \emph{The Twelfth International Conference on Learning Representations}.

\bibitem[{Vaswani et~al.(2017)Vaswani, Shazeer, Parmar, Uszkoreit, Jones, Gomez, Kaiser, and Polosukhin}]{vaswani-etal-2017-transformer}
Ashish Vaswani, Noam Shazeer, Niki Parmar, Jakob Uszkoreit, Llion Jones, Aidan~N Gomez, \L~ukasz Kaiser, and Illia Polosukhin. 2017.
\newblock \href {https://proceedings.neurips.cc/paper_files/paper/2017/file/3f5ee243547dee91fbd053c1c4a845aa-Paper.pdf} {Attention is all you need}.
\newblock In \emph{Advances in Neural Information Processing Systems}, volume~30. Curran Associates, Inc.

\bibitem[{Wang et~al.(2018)Wang, Zhang, and Chen}]{wang-wtal-2018-semiar}
Chunqi Wang, Ji~Zhang, and Haiqing Chen. 2018.
\newblock \href {https://doi.org/10.18653/v1/D18-1044} {Semi-autoregressive neural machine translation}.
\newblock In \emph{Proceedings of the 2018 Conference on Empirical Methods in Natural Language Processing}, pages 479--488, Brussels, Belgium. Association for Computational Linguistics.

\bibitem[{Yao et~al.(2024)Yao, Huang, Liu, and Guo}]{yao-etal-2024-deterministic}
Jiashu Yao, Heyan Huang, Zeming Liu, and Yuhang Guo. 2024.
\newblock \href {https://doi.org/10.18653/v1/2024.findings-acl.481} {Deterministic reversible data augmentation for neural machine translation}.
\newblock In \emph{Findings of the Association for Computational Linguistics: ACL 2024}, pages 8075--8089, Bangkok, Thailand. Association for Computational Linguistics.

\bibitem[{Yu et~al.(2025)Yu, Zhao, Zhu, Xu, Zhou, and Zong}]{yu-etal-2025-simulpl}
Donglei Yu, Yang Zhao, Jie Zhu, Yangyifan Xu, Yu~Zhou, and Chengqing Zong. 2025.
\newblock \href {https://openreview.net/forum?id=XBF63bHDZw} {Simul{PL}: Aligning human preferences in simultaneous machine translation}.
\newblock In \emph{The Thirteenth International Conference on Learning Representations}.

\bibitem[{Yu et~al.(2022)Yu, Li, Koh, Zhang, Pang, Qin, Ku, Xu, Baldridge, and Wu}]{yu-etal-2022-vitvqgan}
Jiahui Yu, Xin Li, Jing~Yu Koh, Han Zhang, Ruoming Pang, James Qin, Alexander Ku, Yuanzhong Xu, Jason Baldridge, and Yonghui Wu. 2022.
\newblock \href {https://openreview.net/forum?id=pfNyExj7z2} {Vector-quantized image modeling with improved {VQGAN}}.
\newblock In \emph{International Conference on Learning Representations}.

\bibitem[{Zeng et~al.(2025)Zeng, Liu, Feng, Huang, and Guo}]{zeng-etal-2025-docmedit}
Li~Zeng, Zeming Liu, Chong Feng, Heyan Huang, and Yuhang Guo. 2025.
\newblock \href {https://doi.org/10.18653/v1/2025.findings-acl.1012} {{D}oc{ME}dit: Towards document-level model editing}.
\newblock In \emph{Findings of the Association for Computational Linguistics: ACL 2025}, pages 19725--19743, Vienna, Austria. Association for Computational Linguistics.

\bibitem[{Zhang et~al.(2024{\natexlab{a}})Zhang, Chen, Wang, Lu, and Qiao}]{zhang-etal-2023-brushtext}
Lingjun Zhang, Xinyuan Chen, Yaohui Wang, Yue Lu, and Yu~Qiao. 2024{\natexlab{a}}.
\newblock \href {https://doi.org/10.1609/aaai.v38i7.28550} {Brush your text: Synthesize any scene text on images via diffusion model}.
\newblock \emph{Proceedings of the AAAI Conference on Artificial Intelligence}, 38(7):7215--7223.

\bibitem[{Zhang et~al.(2018)Zhang, Isola, Efros, Shechtman, and Wang}]{zhang-etal-2018-perceptual}
Richard Zhang, Phillip Isola, Alexei~A. Efros, Eli Shechtman, and Oliver Wang. 2018.
\newblock The unreasonable effectiveness of deep features as a perceptual metric.
\newblock In \emph{Proceedings of the IEEE Conference on Computer Vision and Pattern Recognition (CVPR)}.

\bibitem[{Zhang et~al.(2024{\natexlab{b}})Zhang, Fang, Guo, Ma, Zhang, and Feng}]{zhang-etal-2024-streamspeech}
Shaolei Zhang, Qingkai Fang, Shoutao Guo, Zhengrui Ma, Min Zhang, and Yang Feng. 2024{\natexlab{b}}.
\newblock \href {https://doi.org/10.18653/v1/2024.acl-long.485} {{S}tream{S}peech: Simultaneous speech-to-speech translation with multi-task learning}.
\newblock In \emph{Proceedings of the 62nd Annual Meeting of the Association for Computational Linguistics (Volume 1: Long Papers)}, pages 8964--8986, Bangkok, Thailand. Association for Computational Linguistics.

\bibitem[{Zhang et~al.(2025{\natexlab{a}})Zhang, Zhang, Liang, Chen, Xiang, Zhao, Zhou, and Zong}]{zhang-etal-2025-qrdit}
Zhiyang Zhang, Yaping Zhang, Yupu Liang, Zhiyuan Chen, Lu~Xiang, Yang Zhao, Yu~Zhou, and Chengqing Zong. 2025{\natexlab{a}}.
\newblock \href {https://doi.org/10.18653/v1/2025.findings-acl.372} {A query-response framework for whole-page complex-layout document image translation with relevant regional concentration}.
\newblock In \emph{Findings of the Association for Computational Linguistics: ACL 2025}, pages 7138--7149, Vienna, Austria. Association for Computational Linguistics.

\bibitem[{Zhang et~al.(2025{\natexlab{b}})Zhang, Zhang, Liang, Ma, Xiang, Zhao, Zhou, and Zong}]{zhang-etal-2025-layoutdit}
Zhiyang Zhang, Yaping Zhang, Yupu Liang, Cong Ma, Lu~Xiang, Yang Zhao, Yu~Zhou, and Chengqing Zong. 2025{\natexlab{b}}.
\newblock \href {https://doi.org/10.1109/TPAMI.2025.3530998} {Understand layout and translate text: Unified feature-conductive end-to-end document image translation}.
\newblock \emph{IEEE Transactions on Pattern Analysis and Machine Intelligence}, pages 1--18.

\bibitem[{Zhu et~al.(2023)Zhu, Li, Lei, and Xiong}]{zhu-etal-2023-peit}
Shaolin Zhu, Shangjie Li, Yikun Lei, and Deyi Xiong. 2023.
\newblock \href {https://doi.org/10.18653/v1/2023.acl-long.751} {{PEIT}: Bridging the modality gap with pre-trained models for end-to-end image translation}.
\newblock In \emph{Proceedings of the 61st Annual Meeting of the Association for Computational Linguistics (Volume 1: Long Papers)}, pages 13433--13447, Toronto, Canada. Association for Computational Linguistics.

\end{thebibliography}


\appendix

\section{Details of PRIM}
\label{sec:appendix-testset}

Our PRIM includes $5$ translation directions (En-Ru, En-Fr, En-Ro, En-De, En-Cs), and each direction contains $340$ images.
Each source image corresponds to $2$ reference translation images, which texts are rendered based on translations obtained from GPT4 and Google Translate respectively. In the experiments, BLEU is evaluated using $2$ references, while the computation of COMET and FID is averaging the results from both references.

Data quality of benchmark is a critical issue \cite{li-etal-2025-homebench, lu-etal-2025-transbench, zeng-etal-2025-docmedit}, therefore we additionally perform evaluation on the PRIM benchmark.
Following \citet{yu-etal-2025-simulpl}, the translation quality is evaluated by reference-free \texttt{wmt22-cometkiwi-da} \footnote{\url{https://huggingface.co/Unbabel/wmt22-cometkiwi-da}}. The evaluation scores for the PRIM dataset across different translation directions, and the human-annotated MTed dev and test sets are used as comparison are shown in Table \ref{tab:prim-comparison}. Evaluation results show that the translation quality of the PRIM dataset matches the level of human annotation.

\begin{table*}[t]
    \centering
    \begin{tabular}{ccccc}
    \Xhline{1.5pt}
            & \textbf{PRIM-Google} & \textbf{PRIM-GPT4} & \textbf{MTed dev (human)} & \textbf{MTed test (human)} \\
    \hline
    En-De   & 0.8271      & 0.8246    & 0.8075           & 0.8102            \\
    En-Fr   & 0.8386      & 0.8359    & 0.8269           & 0.8188            \\
    En-Cs   & 0.8388      & 0.8397    & 0.8122           & 0.8102            \\
    En-Ru   & 0.8310      & 0.8308    & 0.7901           & 0.7859            \\
    En-Ro   & 0.8392      & 0.8459    & 0.8280           & 0.8250            \\
    \Xhline{1.5pt}
    \end{tabular}
    \caption{Comparison of translation quality between the PRIM dataset and human-annotated datasets, indicating that the translation quality of the PRIM dataset matches the level of human annotation.}
    \label{tab:prim-comparison}
\end{table*}

\begin{figure*}[htbp]
    \centering
    \includegraphics[width=0.95\linewidth]{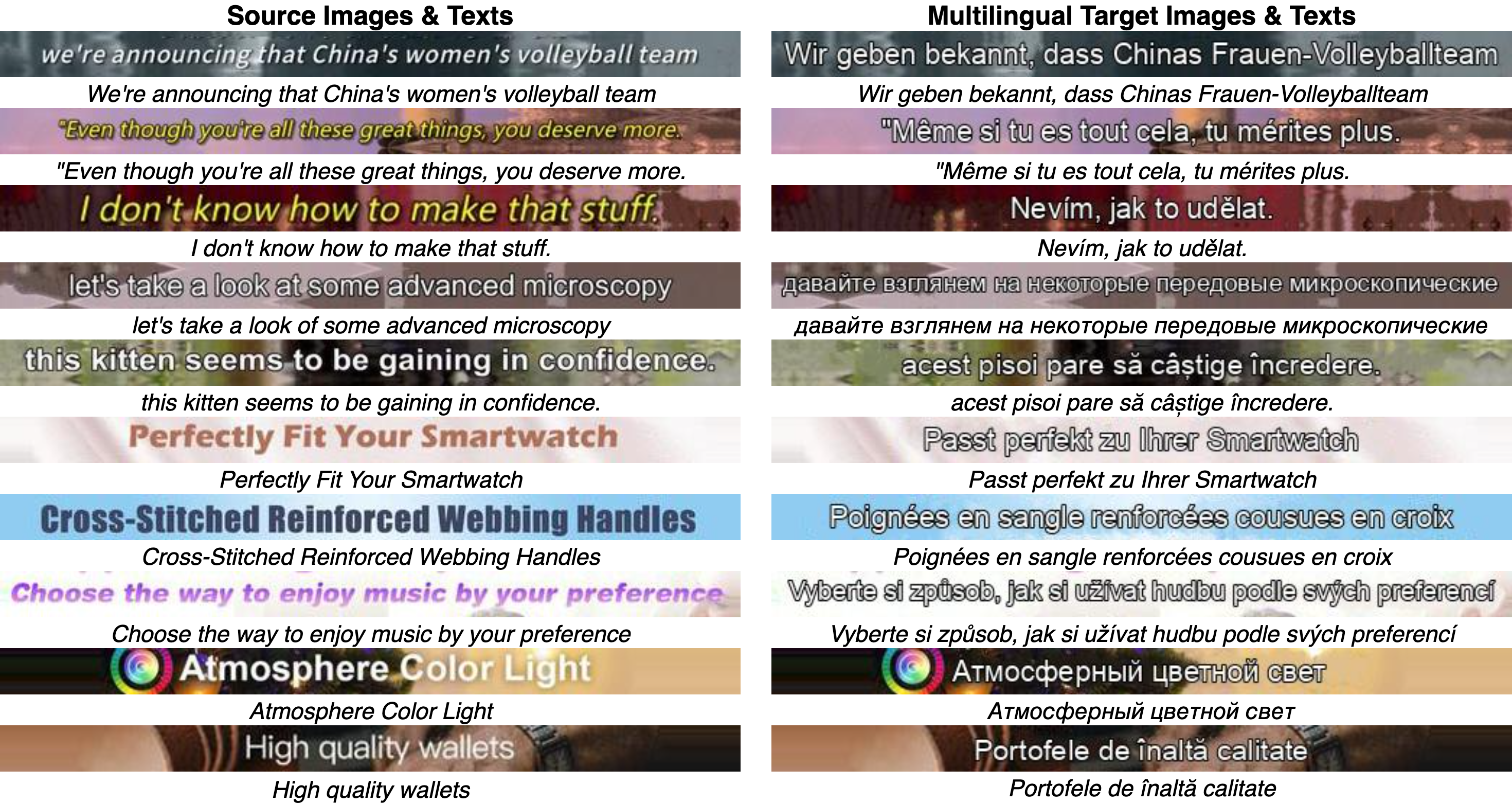}
    \caption{Samples from PRIM, which includes source images, source texts, multilingual target images, and multilingual target texts. The source images are captured from real world, with real-world backgrounds, various fonts, diverse text positions, and $5$ translation directions.}
    \label{fig:test-sample}
\end{figure*}

PRIM only includes source images, target images, source texts, and target texts for evaluation. 
As shown in Figure \ref{fig:test-sample}, the real-world source images demonstrate significant diversity in visual characteristics such as various fonts, diverse text positions.

\section{Details of Training Set}
\label{sec:appendix-trainingset}
The parallel texts used to construct the training set are sourced from the MTed dataset \cite{duh-2018-mted}, which contains transcription texts from Ted talk videos along with multilingual translation results.

Due to texts in MTed dataset are merged, resulting in excessive length unsuitable for rendering into images, we do not directly use the text from the dataset.
Instead, we extract the unmerged texts, along with the corresponding video-related information and transcript timestamps from the original XML documents \footnote{The XML documents are obtained from \url{https://wit3.fbk.eu/}.}.
The source and target language texts are filtered based on timestamps, retaining only parallel texts that can be aligned.

We construct a validation set along with the training set, which is used to evaluate the performance of the model during training. The statistical data for each translation direction is shown in Table \ref{tab:trainingset}.

\begin{table}[htbp]
    \centering
    \begin{tabular}{ccc}
    \Xhline{1.5pt}
    \textbf{Direction} & \textbf{\# Training} & \textbf{\# Validation} \\
    \hline
    En-Ru & 1,629,790 & 3,404 \\
    En-Fr & 1,594,303 & 3,434\\
    En-Ro & 1,507,993 & 3,544 \\
    En-De & 1,418,009 & 3,424 \\
    En-Cs & 848,894 & 3,555\\
    \Xhline{1.5pt}
    \end{tabular}
    \caption{Statistic of training set.}
    \label{tab:trainingset}
\end{table}

Moreover, due to the length differences between source and target language texts with same meaning, the space occupied in the images are also varies. Therefore, the target images adjusts the font size according to the text length. 
Compared to existing publicly available training data, our dataset offers the most diverse styles and translation directions.

\begin{figure*}[htbp]
    \centering
    \includegraphics[width=0.95\linewidth]{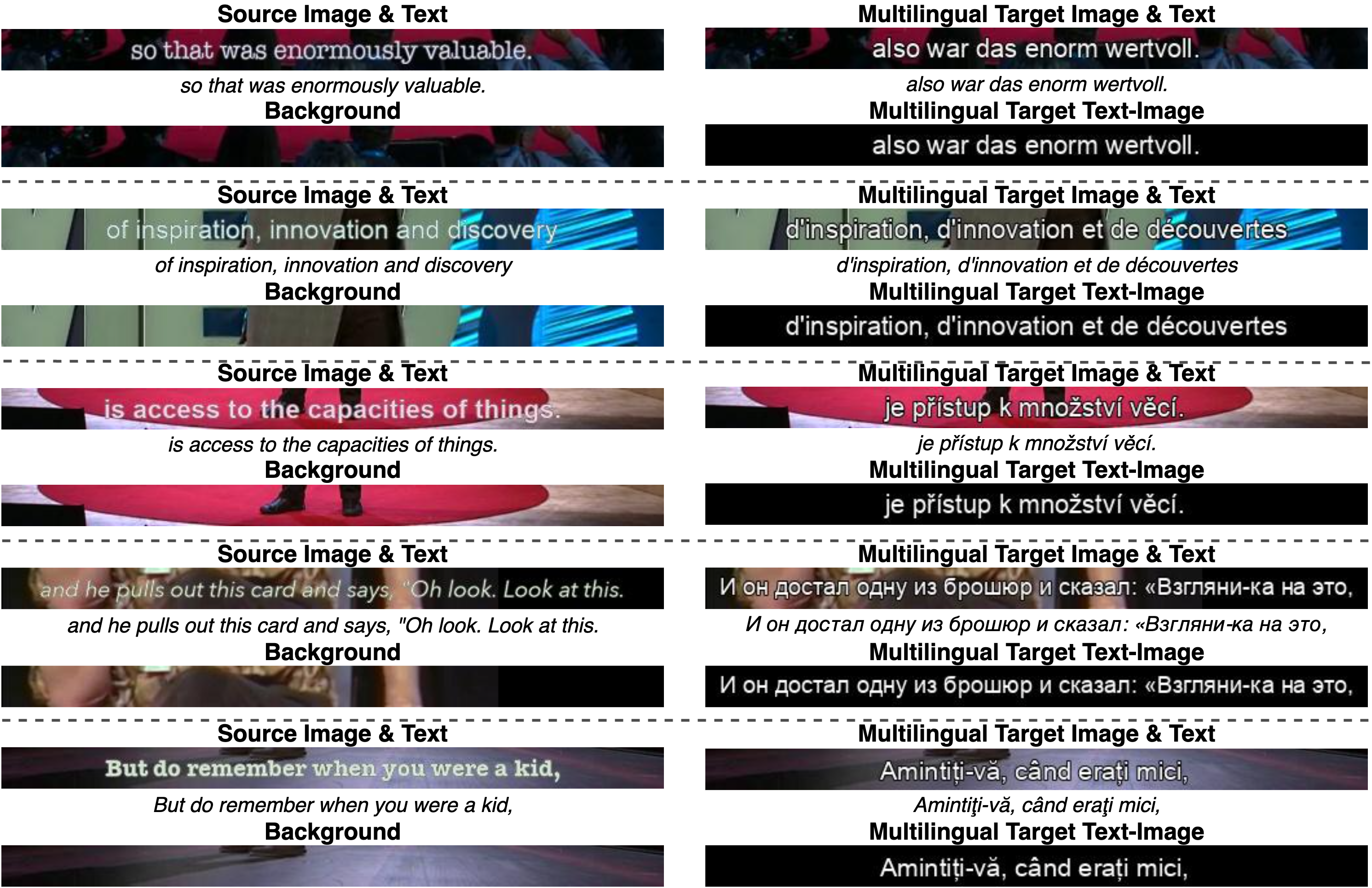}
    \caption{Samples from our training set. Due to the auxiliary training tasks in the training process, the training set includes not only source images, source texts, multilingual target images, multilingual target texts, but also backgrounds and multilingual target text-images.}
    \label{fig:train-sample}
\end{figure*}

Figure \ref{fig:train-sample} shows samples of the training set, including source images, backgrounds, target images, target text-images, source texts, and target text, which enable the two-stage training of our VisTrans model.

\section{Formal Representation of Mask for SAT}
In AT, attention is strictly unidirectional, while the SAT allows bidirectional attention among tokens within the same group. 
The relaxed causal mask $M \in \mathbb{R} ^{n \times n}$ for the sequence length $n$ and group size $K$ can be formalized as follows:
\begin{equation}
M[i][j] =
\begin{cases}
    1, & \text{if } j < \left( [ \frac{i - 1}{K} ] + 1 \right) \times K \\
    0, & \text{other}
\end{cases}
\end{equation}

\section{Implementation of VisTrans}
\label{sec:appendix-implementation-vistrans}
Our VisTrans is trained by Huggingface Accelerate framework \footnote{\url{https://github.com/huggingface/accelerate}} \cite{accelerate} with fp16 mixed precision on $4$ TITAN RTX GPUs.
The implementation of Vision Transformer in VisTrans is referred to timm \footnote{\url{https://github.com/huggingface/pytorch-image-models}}, and the codebook is implemented based on vector-quantize-pytorch \footnote{\url{https://github.com/lucidrains/vector-quantize-pytorch}}.
The texts are tokenized by Sentencepiece \footnote{\url{https://github.com/google/sentencepiece}}.
The perceptual loss is implemented by PerceptualSimilarity \footnote{\url{https://github.com/richzhang/PerceptualSimilarity}}.
Both of two training stages use AdamW optimizer \cite{loshchilov-etal-2018-adamw} with inverse square root learning rate schedule.

The hyperparameters of VisTrans are shown in Table \ref{tab:parameter}, and we choose these parameters based on the performance of model on the validation set.

\begin{table}[htbp]
    \centering
    \begin{tabular}{ccc}
    \Xhline{1.5pt}
    \multirow{5}{*}{\makecell{BackEncoder \\ BackDecoder1 \\ BackDecoder2}} & patch\_size & 16 \\
    & d\_model & 512 \\
    & d\_ff & 2,048 \\
    & heads & 8\\
    & l & 6 \\
    \hline
    \multirow{5}{*}{\makecell{CodeEncoder \\ CodeDecoder1 \\ CodeDecoder2}} & patch\_size & 16 \\
    & d\_model & 512 \\
    & d\_ff & 2,048 \\
    & heads & 8\\
    & l & 6 \\
    \hline
    \multirow{2}{*}{Codebook} & dim & 32 \\
    & size & 8,192 \\
    \hline
    \multirow{5}{*}{ImgDecoder} & patch\_size & 16 \\
    & d\_model & 512 \\
    & d\_ff & 2,048 \\
    & heads & 8\\
    & l & 6 \\
    \hline
    \multirow{5}{*}{MTEncoder} & patch\_size & 8 \\
    & d\_model & 512 \\
    & d\_ff & 2,048 \\
    & heads & 8\\
    & l & 6 \\
    \hline
    \multirow{5}{*}{\makecell{OCR Decoder \\ 1-Pass Decoder}} & d\_model & 512 \\
    & d\_ff & 2,048 \\
    & heads & 8\\
    & l & 6 \\
    & vocabulary & 35,000 \\
    \hline
    \multirow{6}{*}{S2C Decoder} & d\_model & 512 \\
    & d\_ff & 2,048 \\
    & heads & 8 \\
    & l & 3 \\
    & K & 2 \\
    & vocabulary & 176 \\
    \hline
    \multirow{4}{*}{Code Decoder} & d\_model & 512 \\
    & d\_ff & 2,048 \\
    & heads & 8 \\
    & l & 6 \\
    \Xhline{1.5pt}
    \end{tabular}
    \caption{Hyperparameters of VisTrans. }
    \label{tab:parameter}
\end{table}

\section{More Outputs of VisTrans}
\label{sec:moreoutputs}
\begin{figure*}[htbp]
    \centering
    \includegraphics[width=0.95\linewidth]{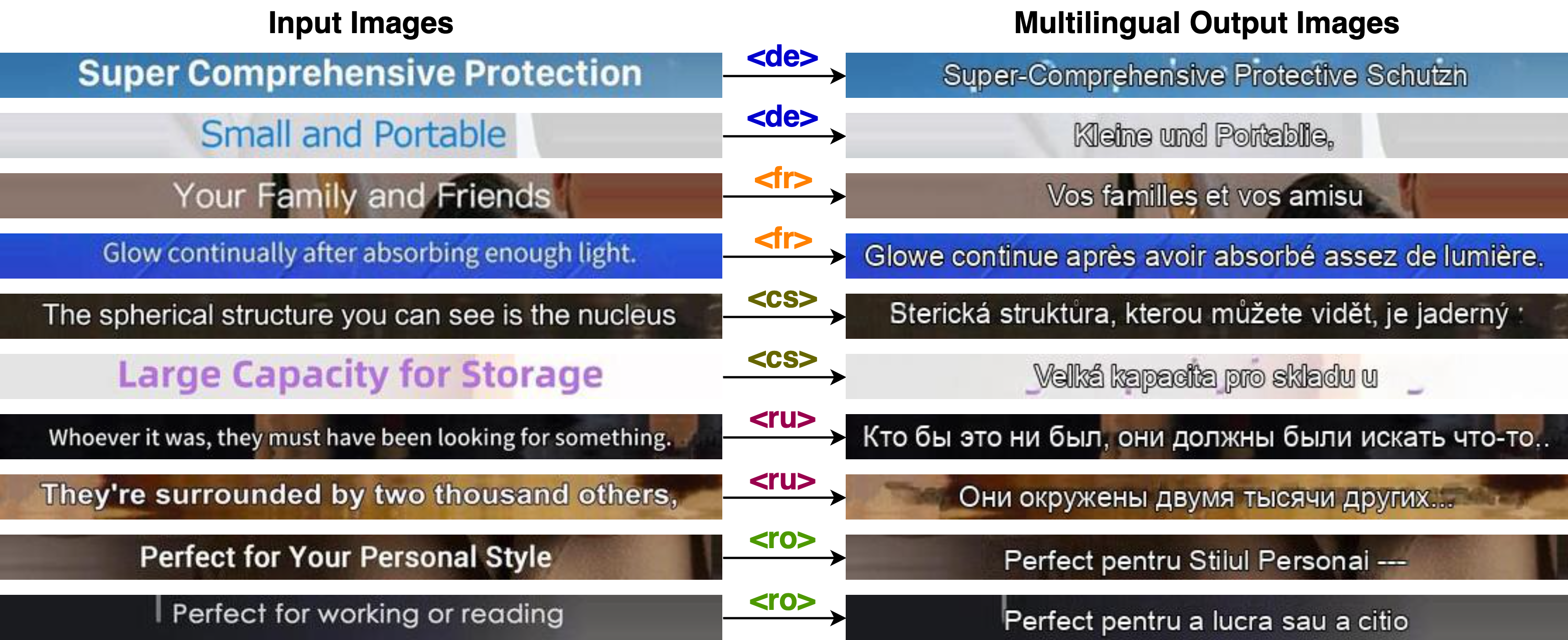}
    \caption{Multilingual output images of VisTrans on PRIM. Our model maintains the integrity of the image background while ensuring multilingual translation performance.}
    \label{fig:more-output}
\end{figure*}

More outputs of our VisTrans for different translation directions are shown in Figure \ref{fig:more-output}.

\section{Explanation of the AnyTrans}
\label{sec:anytrans}
We implement the AnyTrans by PPOCR \footnote{\url{https://github.com/PaddlePaddle/PaddleOCR}}, Qwen1.5-7B \cite{bai-etal-2023-qwen} and AnyText \cite{tuo-etal-2023-anytext}.
The PPOCR firstly detects the text regions and recognizes source texts.
Then, the source texts are translated by Qwen1.5-7B with system prompt \textit{``You are a multilingual translation assistant, and only need to output the translated text.''}, and each source text is add the prompt \textit{``Translate the following text from English to \{target\_language\}: \{source\_text\}''}.

Since AnyText does not support text editing of images with size $32 \times 512$, therefore we resize the images of the test set into $64 \times 512$.
The AnyText requires the original image, the image with text regions removed and the texts prompt as inputs. The text regions in the image are removed based on the regions detected by PPOCR, with regions expanded by a certain proportion. 

\begin{figure}[htbp]
    \centering
    \includegraphics[width=0.95\linewidth]{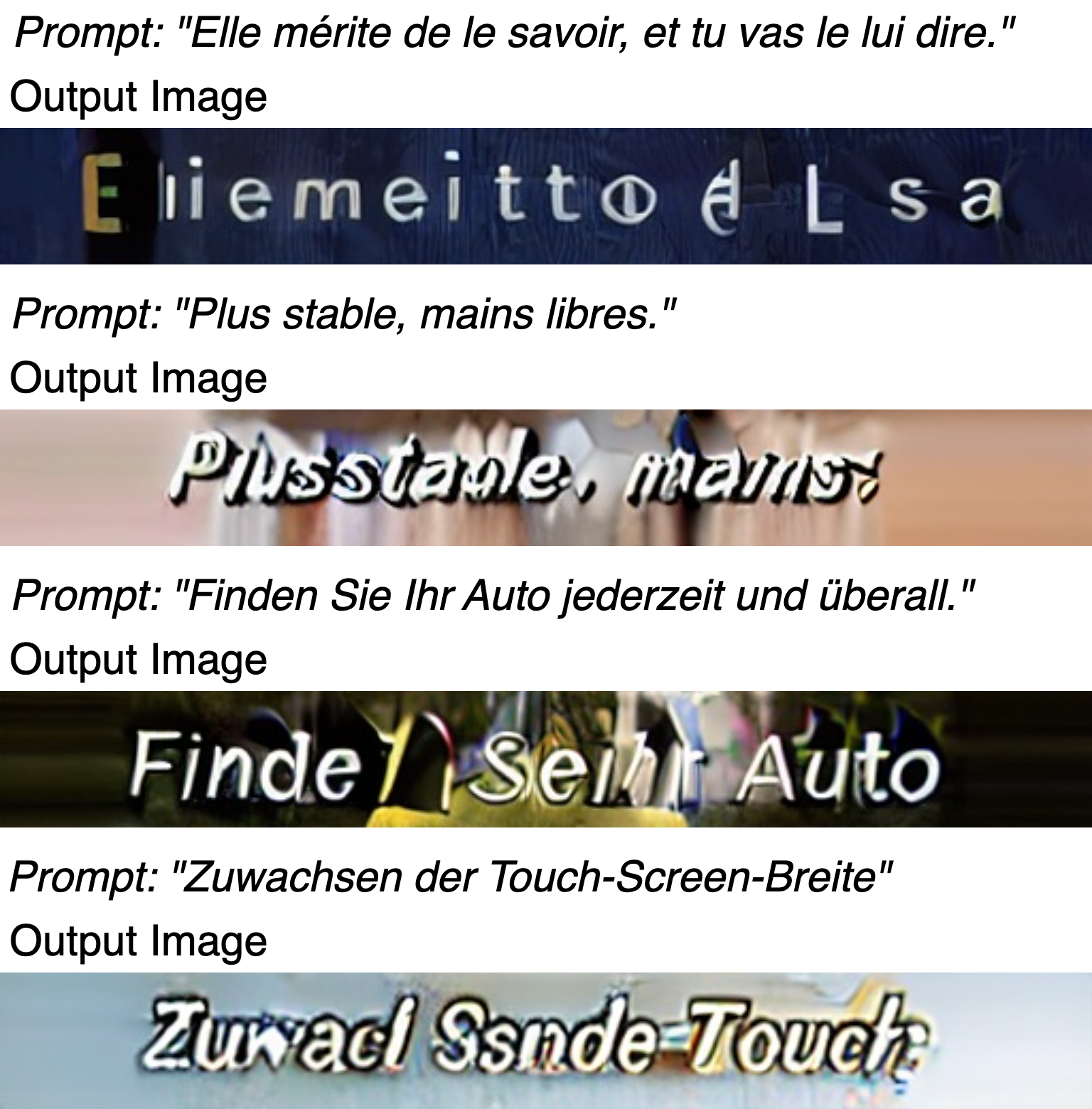}
    \caption{Outputs from AnyTrans \cite{qian-etal-2024-anytrans}. The prompt is the translation result of recognized text by PPOCR, based on Qwen-1.5 7B. We find that AnyTrans is not well-suited for our test set, which is limited by the text editing capability of lengthy text in the image.}
    \label{fig:anytrans-output}
\end{figure}

However, we find that AnyText cannot generate good text editing results in our test set, due to the lengthy text in the images, which occupy a large amount of space. 
Although AnyTrans exhibits strong performance, it is limited by the performance of AnyText, and it is not well-suited for our test set.
We present some outputs from the AnyTrans in Figure \ref{fig:anytrans-output}.

\end{document}